\pdfoutput=1
\documentclass[11pt]{article}

\usepackage[final]{acl}

\usepackage{times}
\usepackage{latexsym}

\usepackage[T1]{fontenc}
\usepackage[utf8]{inputenc}

\usepackage{microtype}

\usepackage{inconsolata}
\usepackage{amsthm}
\newtheorem{proposition}{Proposition}
\usepackage{graphicx}
\usepackage{xcolor}
\usepackage{mdframed}
\usepackage[most]{tcolorbox}
\usepackage{subcaption} % 用于插入子图
\usepackage{hyperref}
\usepackage{booktabs}
\usepackage{algorithm}
\usepackage[noend]{algorithmic}
\usepackage{amssymb}
\usepackage{CJKutf8}
\usepackage{colortbl}
\usepackage{multirow} 
\usepackage{fontawesome5}
\usepackage{academicons}
\definecolor{teaserblue}{RGB}{242, 242, 255}
\definecolor{softpink}{RGB}{255, 230, 240}
\definecolor{pastelgreen}{RGB}{230, 255, 230}
\definecolor{skyblue}{RGB}{230, 240, 255}
\definecolor{warmyellow}{RGB}{255, 250, 230}
\definecolor{softred}{RGB}{255, 230, 230}
\definecolor{deepgreen}{RGB}{200, 255, 200}

\title{Addressing Tokenization Inconsistency in Steganography and Watermarking  Based on Large Language Models}

\author{Ruiyi Yan \and Yugo Murawaki \\
        Graduate School of Informatics, Kyoto University \\ \texttt{ruiyi@nlp.ist.i.kyoto-u.ac.jp}, \texttt{murawaki@i.kyoto-u.ac.jp}}

\begin{document}
\maketitle
\begin{abstract}
%background
% \textit{Glitch tokens}, unexpected tokens improperly segmented by tokenizers from texts, have been highlighted in recent research.
% A sort of anomalous tokens which are named ``glitch tokens'' in large language models (LLMs) have been reported in recent research.
% These tokens can mislead large language models (LLMs) into generating incorrect, irrelevant, or even harmful answers, thus undermining reliability and safety.
Large language models have significantly enhanced the capacities and efficiency of text generation. On the one hand, they have improved the quality of text-based \textit{steganography}. 
On the other hand, they have also underscored the importance of \textit{watermarking}  as a safeguard against malicious misuse.
%Challenge %Methods
In this study, we focus on tokenization inconsistency (TI) between the sender and the receiver in steganography and watermarking, where TI can undermine robustness.
Our investigation reveals that the problematic tokens responsible for TI exhibit two key characteristics: \textbf{infrequency} and \textbf{temporariness}.
Based on these findings, we propose two tailored solutions for TI elimination: a \textit{stepwise verification} method for steganography and a \textit{post-hoc rollback} method for watermarking.
% (1) For LLM-based steganography, we rethink the problem called ``\textit{segmentation ambiguity}'' that was previously solved from a perspective on raw texts. Instead, we solve this problem by addressing glitch tokens, to ensure 100\% accuracy of steganographic extraction, while minimizing overheads of disambiguation.
% (2) For watermarking for LLMs, we investigate the effects of glitch tokens and propose a variant method to address them.
%Experiments
Experiments show that (1) compared to traditional disambiguation methods in steganography, directly addressing TI leads to improvements in fluency, imperceptibility, and anti-steganalysis capacity; (2) for watermarking, addressing TI enhances detectability and robustness against attacks. 
The code is available at \faGithub~\href{https://github.com/ryehr/Consistency}{https://github.com/ryehr/Consistency}.

\end{abstract}

\section{Introduction}

Large language models (LLMs), such as GPT-3~\cite{NEURIPS2020_1457c0d6}, GPT-4~\cite{achiam2023gpt}, Gemini~\cite{team2023gemini,team2024gemini}, and Claude 3~\cite{Claude3}, have revolutionized natural language processing and showcased impressive near-human-level text generation capabilities. 
These advanced LLMs facilitate the creation of flexible and contextually coherent text across diverse genres for text-based \textit{steganography}~\cite{8470163,ziegler-etal-2019-neural}: a promising field in safeguarding information, referring to the art of concealing messages within texts.

However, the same human-like text generation capabilities also pose risks, as synthesized content can be exploited for malicious purposes~\cite{bergman-etal-2022-guiding, MIRSKY2023103006}. To address this, \textit{watermarking} techniques for LLMs~\cite{pmlr-v202-kirchenbauer23a, zhao2024provable} have been developed, embedding imperceptible yet algorithmically detectable signals into generated text. These techniques play a crucial role in ensuring the detectability and responsible use of LLM-generated content.

\begin{figure*}[!t]
    \centering
    % 子图 (a)
    \begin{subfigure}[b]{0.48\textwidth} % 子图宽度为总宽度的45%
        \centering
        \includegraphics[width=\textwidth]{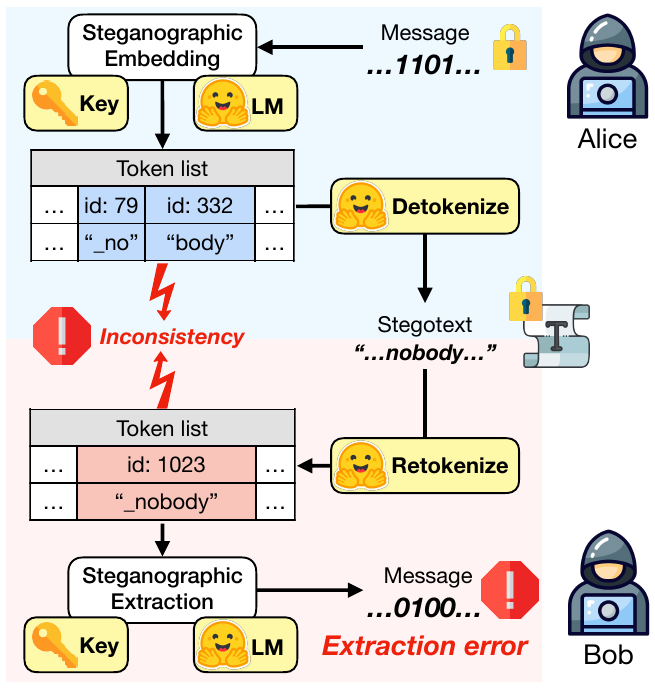} % 替换为你的图片路径
        \caption{Tokenization inconsistency (TI) in steganography.}
        \label{fig: glitch_token_steganography} % 子图标签
    \end{subfigure}
    \hfill % 添加间距
    % 子图 (b)
    \begin{subfigure}[b]{0.48\textwidth}
        \centering
        \includegraphics[width=\textwidth]{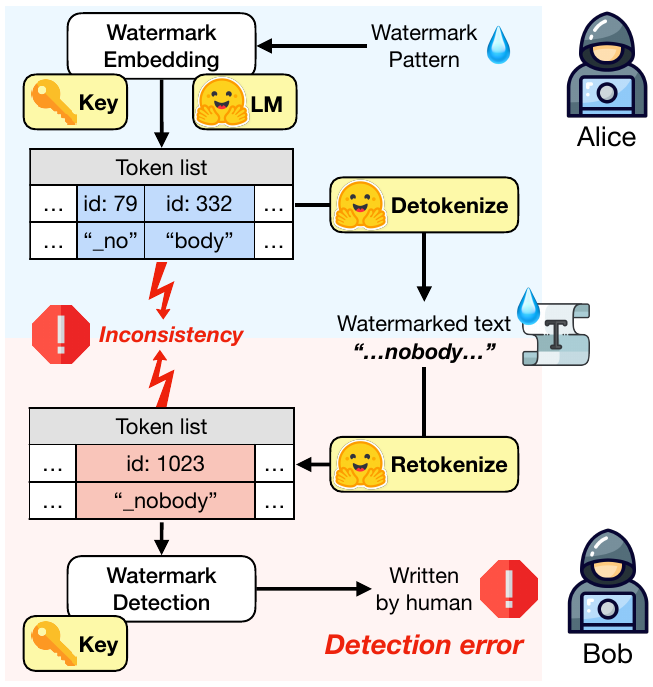} % 替换为你的图片路径
        \caption{Tokenization inconsistency (TI) in watermarking.}
        \label{fig: glitch_token_watermark} % 子图标签
    \end{subfigure}
    % 主图说明
    \caption{An example of tokenization inconsistency (TI) in LLM-based steganography or LLM-based watermarking. Alice generates a token sequence corresponding to subwords ``\_no'' and ``body'' (SITs) during steganography or watermark embedding. During transmission, the generated tokens are detokenized into the text ``nobody''. However, the receiver Bob retokenizes the text ``nobody'' as a single token ``\_nobody'' (a CIT). This can lead to errors in steganography extraction or watermark detection.}
    \label{fig: glitch_token_main} % 主图标签
\end{figure*}

% Glitch tokens refer to a class of anomalous tokens in LLMs that can trigger unexpected and often erroneous behaviors when processed by LLMs. This issue arises from improper tokenization of raw texts, which can stem from irregularities in the training process, such as underrepresentation in training data or inconsistencies in tokenization~\cite{land2024fishing,geiping2024coercing}. 
% Tokenization stands as a cornerstone in natural language processing, which transforms a continuous text sequence into a list of discrete values called tokens~\cite{wang2024tokenization}.
% Previous research~\cite{10.1145/3660799,10.1145/3691620.3695060} has explored the impact of glitch tokens, but their  countermeasures are limited to mitigation. 

% Inspired by the previous definition of glitch tokens in LLMs, we observe that glitch tokens can also occur in LLM-based steganography (a technique embedding secret messages into texts for covert communication) and LLM-based watermarking~\cite{pmlr-v202-kirchenbauer23a} (a technique embedding a hidden pattern into texts to be identified as AI-generated texts).
In both steganography and watermarking applications, Alice (the sender) employs LLMs to generate steganographic texts (stegotexts) or watermarked texts, which are then transmitted to Bob (the receiver).
During this process, an intermediate \textbf{detokenization-retokenization} pipeline is applied to the text as it moves from Alice to Bob. As a result, tokenization inconsistency (TI)~\cite{sun-etal-2023-tokenization} can arise, where discrepancies occur between the originally generated token list and the retokenized token list, potentially impacting the robustness of the system.
Specifically, the inconsistent tokens generated by Alice which are responsible for TI are referred to source inconsistent tokens (SITs), while the corresponding inconsistent tokens resulting from Bob’s retokenization are termed consequential inconsistent tokens (CITs).
% \footnote{Within the detokenization-retokenization pipeline, ~\citet{land2024fishing} can define SIT as unreachable tokens.}
Figure~\ref{fig: glitch_token_main} exemplifies how TI causes negative impacts on steganography (\ref{fig: glitch_token_steganography}) and watermarking (\ref{fig: glitch_token_watermark}). 
% The core problem of glitch tokens in steganography and watermarking is the tokenization inconsistencies between Alice and Bob. 

Inconsistent tokens 
% \begin{mdframed}[backgroundcolor=warmyellow,hidealllines=true]
% This work aims at 100\% addressing glitch tokens for LLM-based steganography and LLM-toward watermarking.
% \end{mdframed}
% LLM-based steganography, a type of generative linguistic steganography and a type of information safeguarding technique, involves utilizing LLMs to conceal messages within text~\cite{ziegler-etal-2019-neural,10.1145/3664647.3680562}. A watermark for LLMs is a hidden pattern in text that is imperceptible to humans, while making the text algorithmically identifiable as synthetic~\cite{pmlr-v202-kirchenbauer23a,hu2024unbiased}.
have not been systematically investigated in view of the detokenization-retokenization pipeline. Especially in steganography and watermarking, they comprise \textbf{robustness}, and can be \textbf{100\% removable}. Besides, any inconsistent token could be catastrophic for most LLM-based steganographic approaches, as any one-step extraction error could cause a series of errors~\cite{qi2024provably}.
Motivated by these facts, this study aims to deepen the understanding of inconsistent tokens in both steganography and watermarking. Specifically, we achieve 100\% correct extraction for steganography with minimal negative impact, and to enhance the detectability and robustness of LLM watermarks. 
The key contributions of this work are as follows: 

\textit{1)} We investigate the emergence of inconsistent tokens during token-by-token generation by language models and identify two key characteristics: \textbf{infrequency} and \textbf{temporariness}.

\textit{2)} Taking advantage of the {infrequency}, we propose a \textit{stepwise verification} method for steganography, maintaining 100\% correct extraction.

\textit{3)} Taking advantage of both the {infrequency} and {temporariness}, we propose a \textit{post-hoc rollback} method for watermarking, which is a lightweight variant method.

\textit{4)} Experiments are conducted across various language models, demonstrating the superiority of our methods: (1) In steganography, compared to the best baseline in each group, our stepwise verification method improves fluency (lowering perplexity by 14.12\%), imperceptibility (lowering KL divergence by 47.86\%), and anti-steganalysis capacity (lowering steganalysis accuracy by 3.53\%) of steganographic texts (stegotexts). (2) For watermarking, our post-hoc rollback method overall enhances the detectability and robustness compared to TI-unaware watermarking.

% An LM generates a sequence of tokens conditioned on a given prompt s. In a single step, during token-by-token generation, we denote the sequence text as $\{t^{(i)} \}^n_{i=1}$, where $t^{(i)}$ represents the $i^{th}$ token in the $n$-token sentences. 
% To generate the next token ($t^{(n+1)}$), the language model predicts the candidate pool of $t^{n+1}$ through $k$ historical tokens (if any) of $Seq$, where $P(t^{n+1}|t^{n-k+1},...,t^{n})$ is the transition probability. The candidate pool for $t_{n+1}$ is:
% \begin{equation}\label{eq: CP}
% {\boldsymbol{\hat{c}}^{(t)}}_{n+1}^o=\{{c}_{n+1}^1,{c}_{n+1}^2,...,{c}_{n+1}^{|{\mathcal{V}}|} \}
% \end{equation}
% with its corresponding logits:
% \begin{equation}\label{eq: logits}
% \mathbf{l}_{n+1}^o=\{{l}_{n+1}^1,{l}_{n+1}^2,...,{l}_{n+1}^{|{\mathcal{V}}|} \} 
% \end{equation}
% with its corresponding probability distribution (using $\mathrm{softmax}$ as an example):
% \begin{equation}\label{eq: PD}
% \boldsymbol{p}^{(t)}_{n+1}^o=\mathrm{softmax}(\mathbf{l}_{n+1}^o)=\{{p}_{n+1}^1,{p}_{n+1}^2,...,{p}_{n+1}^{|{\mathcal{V}}|} \} 
% \end{equation}
% where ${\mathcal{V}}$ is the whole vocabulary of $\mathcal{M}^o$. Then, $t^{n+1}$ is sampled from ${\boldsymbol{\hat{c}}^{(t)}}_{n+1}^o$ according to $\boldsymbol{p}^{(t)}_{n+1}^o$
% % i.e. a finite set of tokens,
% % and $\mathcal{V}_{j=1}^{|{\mathcal{V}}|} p_{n+1}^j=1$.

\section{Investigation: Inconsistent Tokens in Generation by Language Models}
\label{sec: investigation}

We investigate how and to what extent inconsistent tokens emerge during token-by-token generation by language models. This investigation serves as a fundamental basis for studying TI in most steganography and watermarking techniques.
First, we refine the definition of SITs and CITs. Letting the token list generated by Alice $\mathcal{H} = [s^{(0)},s^{(1)},\ldots,s^{(m)}]$, the token list after the detokenization-retokenization pipeline (used by Bob) is $\mathcal{H}^{\prime} = [s^{\prime(0)},s^{\prime(1)},\ldots,s^{\prime(k)}]$. 
The index set of SITs in $\mathcal{H}$ is $I_{\text{SIT}}$ and the index set of CITs in $\mathcal{H}^{\prime}$ is $I_{\text{CIT}}$.
Both $\mathcal{H}$ and $\mathcal{H}^{\prime}$ can be considered to be split from a string $\text{str}$ that consists of $n$ characters. For $\mathcal{H}$, $s^{(i)}$ covers $\text{str}[a_i,b_i]$, and for $\mathcal{H}^{\prime}$, $s^{\prime(j)}$ covers $\text{str}[c_j,d_j]$. Then, $I_{\text{SIT}} = \{ i|[a_i,b_i] \notin \{[c_j,d_j] : j\} \}$ and $I_{\text{CIT}} = \{ j|[c_j,d_j] \notin \{[a_i,b_i] : i\} \}$.

\begin{figure*}[!t]
 \centering
 \includegraphics[width=\textwidth]{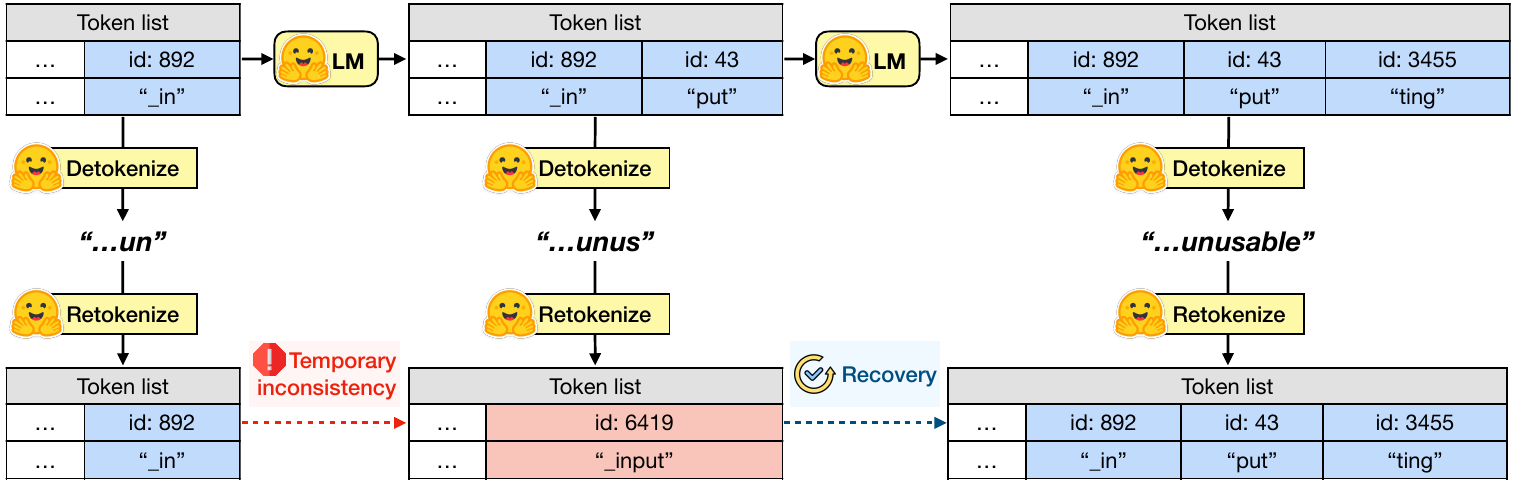} 
 \caption{An example where a candidate-level IT causes TI which recovers back to consistency during token-by-token generation. As the token ``put'' (id: 43) can cause TI immediately after it is output, this token is a candidate-level IT. However, this token does not cause a long-term TI, as TI recovers afterwards.}
 \label{fig: temporary_glitch}
\end{figure*}

We employ three language models: Llama-2-7b\footnote{\href{https://huggingface.co/meta-llama/Llama-2-7b-hf}{https://huggingface.co/meta-llama/Llama-2-7b-hf}}~\cite{touvron2023llama}, Swallow-7b\footnote{\href{https://huggingface.co/tokyotech-llm/Swallow-7b-hf}{https://huggingface.co/tokyotech-llm/Swallow-7b-hf}}~\cite{Fujii:COLM2024,Okazaki:COLM2024} and 
Qwen2.5-7b\footnote{\href{https://huggingface.co/Qwen/Qwen2.5-7B}{https://huggingface.co/Qwen/Qwen2.5-7B}}~\cite{qwen2.5,qwen2}, respectively with English, Japanese and Chinese contexts, to investigate the behavior of inconsistent tokens. Their tokenizers are all based on subwords.
Specifically, the tokenizer of Llama-2-7b is based on BPE (Byte Pair Encoding)~\cite{sennrich2015neural} and on SentencePiece~\cite{kudo2018sentencepiece}, while the tokenizer of Swallow-7b and the tokenizer of Qwen2.5-7b are based on BPE with byte-level fallback. Both Swallow-7b and Qwen2.5-7b possess multilingual capabilities.

For each language model and for each specified number of generated tokens, we generate 1,000 text samples. Texts are produced token by token, with each token sampled using multinomial sampling (in single-track generation).
The experimental setups for this section are detailed in Appendix~\ref{sec:overall_setup}.

\begin{table}[!t]
\renewcommand{\arraystretch}{1.0}
\centering
\small

 \scalebox{1.0}{
\begin{tabular}{l|rrr}
\toprule[1pt]
\begin{tabular}[l]{@{}l@{}}\textbf{Token}\\ \textbf{number}\end{tabular}   & \begin{tabular}[r]{@{}r@{}}\textbf{Llama}\textbf{-2-7b}\end{tabular} & \begin{tabular}[r]{@{}r@{}}\textbf{Swallow}\textbf{-7b}\end{tabular} & \begin{tabular}[r]{@{}r@{}}\textbf{Qwen2.5}\textbf{-7b}\end{tabular} \\ \midrule[1pt]
\textbf{25}  &    3.9\%        &    3.9\%        &    8.7\%        \\
\textbf{50}  &    8.1\%       &    5.0\%        &     11.1\%       \\
\textbf{100} &    17.7\%      &     9.6\%       &      18.1\%      \\
\textbf{200} &    33.6\%       &    15.5\%        &     39.4\%       \\
\textbf{400} &    53.0\%       &    34.2\%        &      66.4\%      \\ \bottomrule[1pt]
\end{tabular}}
 \caption{\textbf{(Text-level inconsistency rates)} Rates of existence of TI.}
 \label{table:text_level_rate}
\end{table}

\begin{table}[!t]
\renewcommand{\arraystretch}{1.0}
\centering
\small

 \scalebox{1.0}{
\begin{tabular}{l|rrr}
\toprule[1pt]
\begin{tabular}[l]{@{}l@{}}\textbf{Token}\\ \textbf{number}\end{tabular}   & \begin{tabular}[r]{@{}r@{}}\textbf{Llama}\textbf{-2-7b}\end{tabular} & \begin{tabular}[r]{@{}r@{}}\textbf{Swallow}\textbf{-7b}\end{tabular} & \begin{tabular}[r]{@{}r@{}}\textbf{Qwen2.5}\textbf{-7b}\end{tabular} \\ \midrule[1pt]
\textbf{25}  &    0.176\%        &   0.242\%        &    0.558\%        \\
\textbf{50}  &    0.204\%       &    0.181\%        &     0.381\%       \\
\textbf{100} &    0.215\%      &     0.185\%       &      0.349\%      \\
\textbf{200} &    0.228\%       &    0.157\%        &     0.400\%       \\
\textbf{400} &    0.223\%       &    0.186\%        &     0.467\%      \\ \bottomrule[1pt]
\end{tabular}}
 \caption{\textbf{(Token-level inconsistency rates)} Ratios of the sum of SITs and CITs to the sum of generated tokens and retokenized tokens.}
  \label{table:token_level_rate}
\end{table}

\textbf{Text-level inconsistency rate:} The rate at which TI appears ($\mathcal{H} \neq \mathcal{H}^{\prime}$).
% Specifically, considering a generated token list $\mathcal{H}$, if $\text{tokenize}(\text{detokenize}(\mathcal{H})) \neq \mathcal{H}$, TI appears.
Table~\ref{table:text_level_rate} presents the text-level inconsistency rates across various language models and token generation lengths. Since each token carries some potential to become an inconsistent token, the general trend indicates that the incidence of inconsistent tokenization increases as the number of generated tokens grows.

\textbf{Token-level inconsistency rate}: The ratio of the sum of SITs and CITs to the sum of generated and retokenized tokens, i.e., $\frac{|I_{\text{SIT}}| + |I_{\text{CIT}}|}{|\mathcal{H}|+|\mathcal{H}^{\prime}|}$.
% Specifically, we compare the differences in token frequencies between generated token lists and retokenized token lists.
According to Table~\ref{table:token_level_rate}, this metric is not closely related to text length. Another notable observation is that token-level inconsistency rates are typically below 0.5\%, which results in much higher text-level inconsistency rates though.
The disparity between high text-level inconsistency rates and low token-level inconsistency rates highlights \textit{the infrequency of inconsistent tokens}.

\textbf{Candidate-level inconsistency rate}: The ratio of candidate tokens that can cause TI to all tokens in candidate pools of token-by-token generative steps. 
% These problematic tokens are referred to candidate-level glitch tokens.
To further explore the cause of the infrequency of inconsistent tokens, potential TI from candidate pools are investigated.
% If appending a token to the generated token list results in tokenization inconsistency after detokenization and retokenization, the token is considered a candidate-level glitch token. 
How to determine if a candidate token is a candidate-level inconsistent token (candidate-level IT) is shown in Algorithm~\ref{algorithm: Identify a candidate-level glitch token}, which refers to a detokenization-retokenization pipeline in one step.
Specifically, considering a candidate pool $\boldsymbol{c}^{(t)} = [\text{token}_1, \text{token}_2, \ldots, \text{token}_{|\mathcal{V}|}]$ at time $t$, and a generated historical token list $\mathcal{H}^{(t-1)}$ after time $t-1$. Let $\mathcal{H}^{(t)} = \mathcal{H}^{(t-1)} || \text{token}_i$. If $\text{tokenize}(\text{detokenize}(\mathcal{H}_i^{(t)})) \neq \mathcal{H}_i^{(t)}$, $\text{token}_i$ is a candidate-level IT at time $t$.
For accurate calculation, those scenarios where TI has occurred before outputting a token are excluded.

\begin{table}[!t]
\renewcommand{\arraystretch}{1.0}
\centering
% \small

 \scalebox{0.8}{
\begin{tabular}{l|rrr}
\toprule[1pt]
 & \begin{tabular}[r]{@{}r@{}}\textbf{Llama-2-7b}\end{tabular} & \begin{tabular}[r]{@{}r@{}}\textbf{Swallow-7b}\end{tabular} & \begin{tabular}[r]{@{}r@{}}\textbf{Qwen2.5-7b}\end{tabular} \\ \midrule[1pt]
\begin{tabular}[l]{@{}l@{}}\textbf{Number}\\ \textbf{ratio}\end{tabular} &   1.497\%        &   2.045\%        &    3.993\%        \\ \hline
\begin{tabular}[l]{@{}l@{}}\textbf{Probability}\\ \textbf{ratio}\end{tabular} & 0.184\% & 1.126\% & 1.833\%  \\ \bottomrule[1pt]
\end{tabular}}
 \caption{\textbf{(Candidate-level inconsistency rates)} Number ratios and probability ratios of candidate-level ITs to tokens in candidate pools.}
 \label{table:candidate_level_rate}
\end{table}

For simplicity, only the 64 highest probability tokens in each candidate pool are considered. 
Table~\ref{table:candidate_level_rate} respectively lists the number ratios and probability ratios of candidate-level ITs (the ratios of the cumulative numbers or probabilities of candidate-level ITs to the cumulative numbers or probabilities of top-64 tokens) across various language models, with data aggregated across various text lengths. The results indicate that the infrequency of inconsistent tokens is primarily due to the low candidate-level inconsistency rates in each candidate pool. The infrequency of the SITs is also revealed.

However, when observing the data in Tables~\ref{table:text_level_rate},~\ref{table:token_level_rate}, and~\ref{table:candidate_level_rate}, another question arises: 
% Compared to Swallow-7b, in Llama-2-7b, how can a relatively lower probability ratio of candidate-level ITs in candidate pools (0.184\% compare to 1.126\%), lead to higher text-level inconsistency rates (17.7\% compared to 9.6\% when token number is 100) and higher token-level inconsistency rates (0.215\% compared to 0.185\% when token number is 100)?
How does Llama-2-7b, despite having a lower candidate-level IT probability ratio in candidate pools (0.184\% vs. 1.126\% in Swallow-7b), exhibit higher inconsistency rates at both the text level (17.7\% vs. 9.6\% with 100 tokens) and the token level (0.215\% vs. 0.185\% with 100 tokens)?
Considering limitations in defining the candidate-level inconsistency rate,  it is likely that generating candidate-level ITs may only result in \textit{temporary} TI.
% In other words, those candidate-level ITs could be not SITs in the final, which may no longer lead to TI after subsequent generations.
Figure~\ref{fig: temporary_glitch} instantiates this phenomenon.

\textbf{Temporary inconsistency rate:} Among the candidate-level ITs that are output, this metric represents the rate of temporary SITs (that do not cause TI after the entire generation process).
Table~\ref{table:temporary_glitch_rate} lists the temporary inconsistency rates for the texts generated by the three language models. The significantly lower temporary inconsistency rate for Llama-2-7b compared to others indicates that its generated potential SITs are more stable in affecting the final tokenization. This stability leads to higher text-level and token-level inconsistency rates than those observed for Swallow-7b. When using default tokenizer parameters, a fair amount of `<s>' and `</s>' output by Llama-2-7b lead to stable TI. Besides, Swallow-7b and Qwen2.5-7b are featured by outputing a fair amount of partial UTF-8 tokens~\cite{land2024fishing}. Appendix~\ref{sec: determine q} provides supplements for it.

% The primary reason for the greater stability of glitch tokens is that the LLaMA tokenizer is a BPE (Byte Pair Encoding) model~\cite{sennrich2015neural} based on SentencePiece~\cite{kudo2018sentencepiece}, while others are not based on SentencePiece.

\begin{table}[!t]
\renewcommand{\arraystretch}{1.0}
\centering
% \small
 \scalebox{1.0}{
\begin{tabular}{ccc}
\toprule[1pt]
  \begin{tabular}[c]{@{}c@{}}\textbf{Llama}\textbf{-2-7b}\end{tabular} & \begin{tabular}[c]{@{}c@{}}\textbf{Swallow}\textbf{-7b}\end{tabular} & \begin{tabular}[c]{@{}c@{}}\textbf{Qwen2.5}\textbf{-7b}\end{tabular} \\ \midrule[1pt]
  8.76\%        &   81.98\%        &    87.93\%        \\  \bottomrule[1pt]
\end{tabular}}
 \caption{\textbf{(Temporary inconsistency rates)} Rates of candidate-level ITs that do not cause TI in final among all candidate-level ITs.}
 \label{table:temporary_glitch_rate}
\end{table}

In summary, our investigation reveals that inconsistent tokens generated by language models are characterized as \textit{(1) infrequency} and \textit{(2) temporariness}. These findings inspire us to develop methods to address inconsistent tokens and TI in LLM-based steganography and watermarking.

\begin{algorithm}[!t]
\small
\caption{Identify a candidate-level IT}\label{algorithm: Identify a candidate-level glitch token} 
\textbf{Input:}\\
Token to be verified, $s_o$ \\
Previously generated token list, $L$  \\
{\textbf{Output:}}\\
Candidate-level IT or not (True or False), $\mathrm{Result_G}$\\

\begin{algorithmic}[1]
\STATE Append $s_o$ to $L$ to obtain $L_o$.
\\\textit{/* Token list to be verified*/}
\STATE Detokenize $L_o$ into a temporary text $t_{temp}$.
\STATE Tokenize $t_{temp}$ into $L'$.
\STATE $\mathrm{Result_G} \leftarrow \neg (L_o == L')$;

\RETURN $\mathrm{Result_G}$
\end{algorithmic}

\end{algorithm}

\section{Methods}

In this section, the introduced methods are targeted to the mechanisms of LLM-based steganography and watermarking respectively, meanwhile taking advantage of the infrequency and temporariness of inconsistent tokens.
Specifically: 

\textbf{For steganography:} We propose a stepwise verification method that precisely removes candidate-level ITs at each generation step. As only outputting candidate-level ITs can cause (at least temporary) TI, the presence of candidate-level ITs is a necessary condition for the eventual occurrence of inconsistent tokens. Detailed analysis and explanations about SITs, CITs, TI and candidate-level ITs are shown in Appendix~\ref{sec: Analysis of Inconsistent Tokens and Tokenization Inconsistency (TI)}.
Therefore, the absence of candidate-level ITs is a sufficient condition for the final absence of inconsistent tokens. Hence, our method eliminates TI in the final output.

\begin{figure*}[!t]
 \centering
 \includegraphics[width=\textwidth]{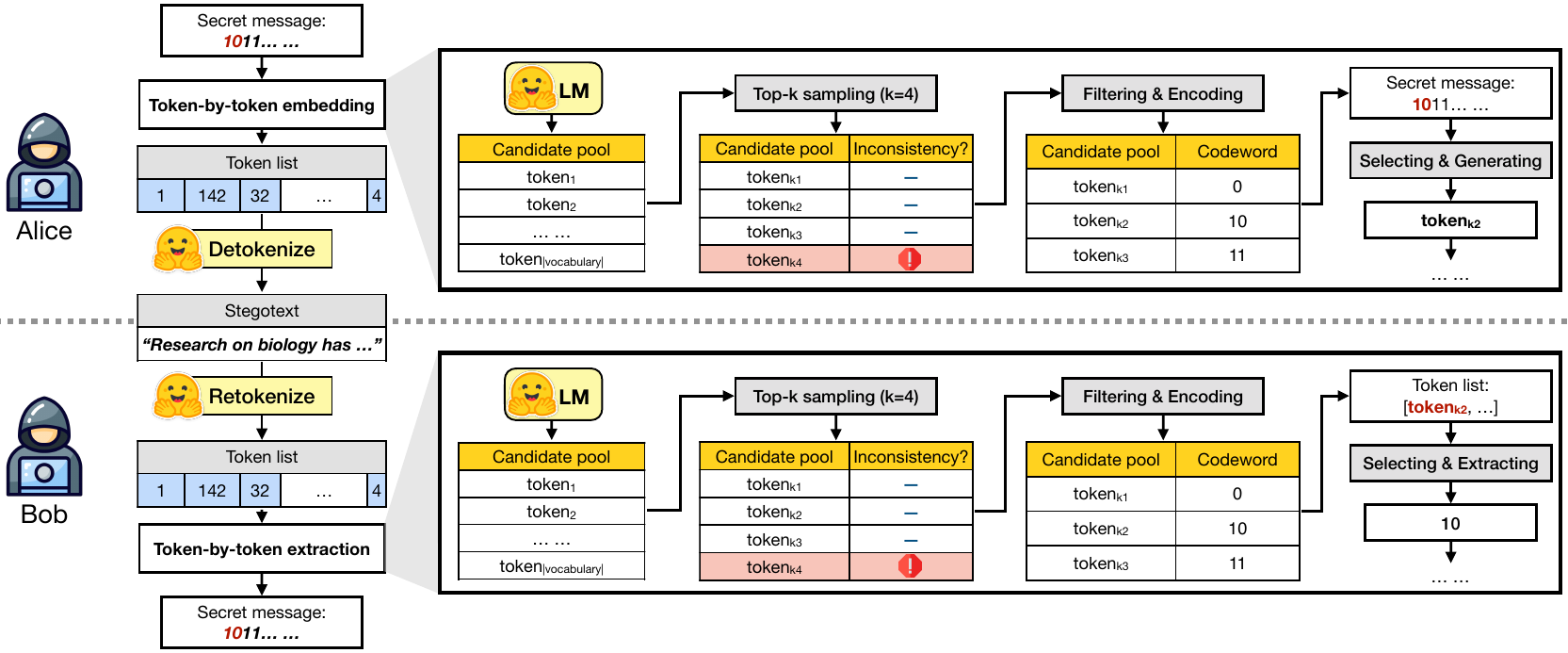} 
 \caption{Overview and procedures of LM-based steganography with our stepwise consistency-verification approach. For some simplicity in this example, Huffman encoding~\cite{8470163} and top-4 sampling in each candidate pool is adopted. The verification mechanism is in place before encoding for both Alice and Bob.}
 \label{fig: steganography_overview}
\end{figure*}

\begin{figure}[!t]
    \centering
    % 子图 (a)
    \begin{subfigure}[b]{\columnwidth} % 子图宽度为总宽度的45%
        \centering
        \includegraphics[width=\columnwidth]{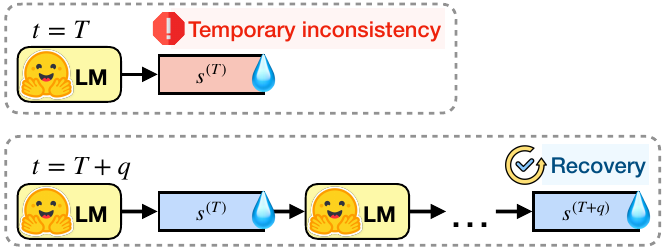} % 替换为你的图片路径
        \caption{Automatically recoverable inconsistency situation.}
        \label{fig: watermark_case_1} % 子图标签
    \end{subfigure}
    \hfill % 添加间距
    % 子图 (b)
    \begin{subfigure}[b]{\columnwidth}
        \centering
        \includegraphics[width=\columnwidth]{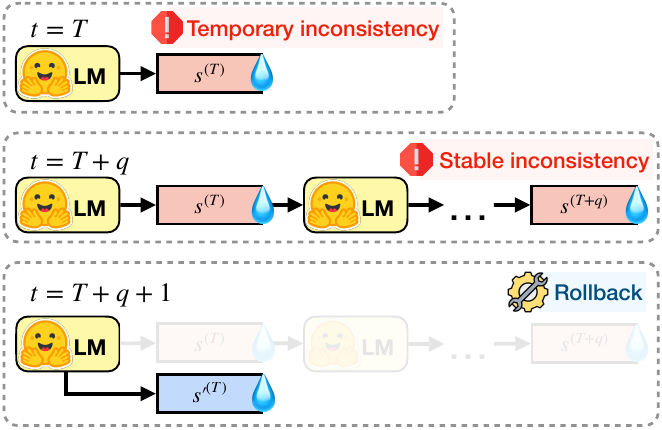} % 替换为你的图片路径
        \caption{Stable inconsistency situation which needs rollback.}
        \label{fig: watermark_case_2} % 子图标签
    \end{subfigure}
    % 主图说明
    \caption{Mechanisms of the post-hoc rollback method. Due to temporariness of inconsistent tokens, a $q$-token observation period is set for them.}
    \label{fig: watermark_cases} % 主图标签
\end{figure}

In steganography, since candidate tokens are associated with codewords, it is necessary to call the tokenizer to verify whether each candidate token is a candidate-level inconsistent token. Although this process may appear inefficient, our method operates with linear complexity, still providing some advantages over the previous disambiguation algorithms~\cite{nozaki-murawaki-2022-addressing,yan2023A,qi2024provably} whose complexities are at least $\mathrm{O}(n^2)$. Both our method for addressing TI and previous disambiguation approaches share the same goal: ensuring 100\% correct steganographic extraction.

Additionally, our method preemptively removes candidate-level ITs to achieve extraction accuracy. Since both candidate-level ITs and final inconsistent tokens are infrequent, KL divergence between the original and the modified candidate pools (resulting from removing candidate tokens) remains small, greatly reducing the negative impact on imperceptibility. Theories on imperceptibility are provided in Appendix~\ref{sec: imperceptibility_GLS}.

\textbf{For watermarking}: We propose a post-hoc rollback method that makes the generation process rollback to the state where TI has not happened if TI persists. The rollback mechanism does not respond immediately to TI because of their temporariness.
Unlike the method used for steganography, there is a higher relaxation for watermarking, because the detector does not require detailed information of candidate pools at each step to decode the text, so that candidate tokens do not need to be examined individually. Negative effects on imperceptibility can be nearly negligible because of the infrequency of the output inconsistent tokens.

\subsection{A Stepwise Verification Method for Steganography}

The overview of the stepwise verification method is shown in Figure~\ref{fig: steganography_overview}, where the verification mechanism is placed between the sampling and steganographic encoding steps. Both the sender and receiver can verify whether each token in the candidate pool is a candidate-level IT, enabling them to perform steganographic encoding on the same filtered candidate pools. This guarantees that the receiver can accurately extract the secret messages.
How to identify a candidate-level IT from a candidate pool is detailed in Algorithm~\ref{algorithm: Identify a candidate-level glitch token}. 
% First, a token in the candidate pool  $s_o$ is appended  to the existing token list  $L$  to form \( L_o\) (Line 1). Next, \(L_o\) is detokenized into a text string $t_{temp}$, and then retokenized into a token list $L'$ (Line 2-3). Finally, whether \( L_o\)  is identical to $L'$ is checked, returning a Boolean result (Line 4-5).

The process simulates detokenization and retokenization of the generated stegotext transmitted from Alice to Bob. candidate-level ITs are removed from the candidate pool, as they could interfere Bob's extraction process. By eliminating such problematic tokens, the approach ensures that both Alice and Bob maintain identical token sequences, enabling reliable steganographic extraction.\footnote{Further details can be found in Appendix~\ref{sec: details_stepwise}.}

\subsection{A Post-Hoc Rollback Method for Watermarking}
This method of removing inconsistent tokens for LLM watermarking leverages not only the infrequency of inconsistent tokens but also their temporariness. The overview of this post-hoc method is illustrated in Figure~\ref{fig: watermark_cases}, which only involves the token-by-token watermark embedding.

The core idea is as follows: If a candidate-level IT is generated and causes temporary TI at that step, the token is assigned an \textit{observation period} that lasts until the next $q$ tokens are generated. Once the observation period ends, if tokenization consistency is recovered (Figure~\ref{fig: watermark_case_1}), no further action is necessary. However, if TI persists, this temporary TI is deemed a stable TI, and the generation process rolls back to the state before the candidate-level IT was generated. At this regressed point, the candidate pool is resampled, excluding that candidate-level IT (Figure~\ref{fig: watermark_case_2}).\footnote{More detailed operation steps for this method for watermarking are shown in  Appendix~\ref{sec: details_posthoc}.}

\begin{table*}[!t]
\small
\renewcommand{\arraystretch}{1.0}
\centering

  \scalebox{1.0}{
\begin{tabular}{l|rrrr|rrrr}
\toprule[1.0pt]
          % \multicolumn{9}{c}{\textbf{Llama2-2-7b}}   \\ \hline
         & \multicolumn{4}{c|}{\textbf{2.0 $\leq$ BPT $<$ 3.0}} & \multicolumn{4}{c}{\textbf{3.0 $\leq$ BPT $<$ 4.0}}  \\ \cline{2-9}
         
         & \textbf{PPL$\downarrow$}  & \textbf{KLD$\downarrow$}   & \textbf{ACC$\downarrow$}  & \textbf{Time$\downarrow$}  & \textbf{PPL$\downarrow$}   & \textbf{KLD$\downarrow$}   & \textbf{ACC$\downarrow$}  & \textbf{Time$\downarrow$}     \\ \hline
\textbf{Basic}    & \cellcolor{softred}17.72 & \cellcolor{softred} 1.005 & \cellcolor{softred} 0.936 & \cellcolor{pastelgreen} 1.62 & \cellcolor{softred} 38.96 & \cellcolor{softred} 1.355 & \cellcolor{softred} 0.976 & \cellcolor{pastelgreen} 1.43 \\
\textbf{MWIS}     & 9.13 & 0.138 & 0.784 & \cellcolor{softred} 6.33 & 16.37 & 0.149 & \cellcolor{pastelgreen} 0.797 & \cellcolor{softred} 11.40 \\
\textbf{SyncPool} & 10.73 & 0.154 & 0.853 & 3.04 & $-$ & $-$ & $-$ & $-$ \\
\textbf{Stepwise verification}     &  \cellcolor{pastelgreen}8.69 & \cellcolor{pastelgreen} 0.070 & \cellcolor{pastelgreen} 0.766 & 5.35 & \cellcolor{pastelgreen} 15.07 & \cellcolor{pastelgreen} 0.031 & 0.846 & 8.09 \\
\midrule[1.0pt]
         & \multicolumn{4}{c|}{\textbf{4.0 $\leq$ BPT $<$ 5.0}} & \multicolumn{4}{c}{\textbf{5.0 $\leq$ BPT $<$ 6.0}}  \\ \cline{2-9}
         & \textbf{PPL$\downarrow$}  & \textbf{KLD$\downarrow$}   & \textbf{ACC$\downarrow$}  & \textbf{Time$\downarrow$}  & \textbf{PPL$\downarrow$}   & \textbf{KLD$\downarrow$}   & \textbf{ACC$\downarrow$}  & \textbf{Time$\downarrow$}  \\ \hline
\textbf{Basic}    & \cellcolor{softred} 77.91 & \cellcolor{softred} 1.690 & \cellcolor{softred} 0.963 & \cellcolor{pastelgreen} 2.73 & \cellcolor{softred}134.87 & \cellcolor{softred}1.934 & \cellcolor{softred}0.922 & \cellcolor{pastelgreen}7.21 \\
\textbf{MWIS}     & 28.64 & 0.187 & 0.625 & \cellcolor{softred} 12.04 & $-$ & $-$ & $-$ & $-$ \\
\textbf{SyncPool} & $-$ & $-$ & $-$ & $-$ & $-$ & $-$ & $-$ & $-$ \\
\textbf{Stepwise verification}     & \cellcolor{pastelgreen}  26.41 & \cellcolor{pastelgreen}  0.022 & \cellcolor{pastelgreen}  0.621 & 9.10 & \cellcolor{pastelgreen}47.73 & \cellcolor{pastelgreen}0.020 & \cellcolor{pastelgreen}0.704 & \cellcolor{softred}9.51 \\
\bottomrule[1.0pt]
\end{tabular}}
 \caption{Quantitative comparison with previous disambiguation methods on Llama-2-7b and English contexts.}
 \label{table: steganography_llama_results}
\end{table*}

\begin{table*}[!t]
\small
\renewcommand{\arraystretch}{1.0}
\centering

  \scalebox{1.0}{
\begin{tabular}{l|rrrr|rrrr}
\toprule[1.0pt]
          % \multicolumn{9}{c}{\textbf{Llama2-2-7b}}   \\ \hline
         & \multicolumn{4}{c|}{\textbf{2.0 $\leq$ BPT $<$ 3.0}} & \multicolumn{4}{c}{\textbf{3.0 $\leq$ BPT $<$ 4.0}}  \\ \cline{2-9}
         
         & \textbf{PPL$\downarrow$}  & \textbf{KLD$\downarrow$}   & \textbf{ACC$\downarrow$}  & \textbf{Time$\downarrow$}  & \textbf{PPL$\downarrow$}   & \textbf{KLD$\downarrow$}   & \textbf{ACC$\downarrow$}  & \textbf{Time$\downarrow$}     \\ \hline
\textbf{Basic}    & \cellcolor{softred}18.18 & \cellcolor{softred}0.834 & \cellcolor{softred}0.952 & \cellcolor{pastelgreen} 1.90 & \cellcolor{softred}36.60 & \cellcolor{softred}0.929 & \cellcolor{softred}0.961 & \cellcolor{pastelgreen} 3.62 \\
\textbf{MWIS}     & 9.49 & 0.126 & 0.866 & \cellcolor{softred}6.36 & 18.70 & 0.132 & 0.915 & \cellcolor{softred}12.12 \\
\textbf{SyncPool} & 12.70 & 0.172 & 0.890 & 3.40 & $-$ & $-$ & $-$ & $-$\\
\textbf{Stepwise verification}     & \cellcolor{pastelgreen} 9.24 & \cellcolor{pastelgreen} 0.081 & \cellcolor{pastelgreen} 0.865 & 4.89 & \cellcolor{pastelgreen}18.03 & \cellcolor{pastelgreen}0.052 & \cellcolor{pastelgreen}0.911 & 7.26 \\
\midrule[1.0pt]
         & \multicolumn{4}{c|}{\textbf{4.0 $\leq$ BPT $<$ 5.0}} & \multicolumn{4}{c}{\textbf{5.0 $\leq$ BPT $<$ 6.0}}  \\ \cline{2-9}
         & \textbf{PPL$\downarrow$}  & \textbf{KLD$\downarrow$}   & \textbf{ACC$\downarrow$}  & \textbf{Time$\downarrow$}  & \textbf{PPL$\downarrow$}   & \textbf{KLD$\downarrow$}   & \textbf{ACC$\downarrow$}  & \textbf{Time$\downarrow$}  \\ \hline
\textbf{Basic}    & \cellcolor{softred}73.11 & \cellcolor{softred}0.995 & \cellcolor{softred}0.984 & \cellcolor{pastelgreen}6.04 & \cellcolor{softred}130.71 & \cellcolor{softred}0.985 & \cellcolor{softred}0.948 & \cellcolor{pastelgreen}7.32 \\
\textbf{MWIS}     & 37.85 & 0.157 & \cellcolor{pastelgreen}0.861 & \cellcolor{softred}13.35 & $-$ & $-$ & $-$ & $-$ \\
\textbf{SyncPool} & $-$ & $-$ & $-$ & $-$ & $-$ & $-$ & $-$ & $-$ \\
\textbf{Stepwise verification}     & \cellcolor{pastelgreen}34.80 & \cellcolor{pastelgreen}0.041 & 0.922 & 8.59 & \cellcolor{pastelgreen}69.81 & \cellcolor{pastelgreen}0.041 & \cellcolor{pastelgreen}0.872 & \cellcolor{softred}9.25 \\
\bottomrule[1.0pt]
\end{tabular}}
 \caption{Quantitative comparison with previous disambiguation methods on Swallow-7b and Japanese contexts.}
 \label{table: steganography_swallow_results}
\end{table*}

\begin{table*}[!t]
\small
\renewcommand{\arraystretch}{1.0}
\centering

  \scalebox{1.0}{
\begin{tabular}{l|rrrr|rrrr}
\toprule[1.0pt]
          % \multicolumn{9}{c}{\textbf{Llama2-2-7b}}   \\ \hline
         & \multicolumn{4}{c|}{\textbf{2.0 $\leq$ BPT $<$ 3.0}} & \multicolumn{4}{c}{\textbf{3.0 $\leq$ BPT $<$ 4.0}}  \\ \cline{2-9}
         
         & \textbf{PPL$\downarrow$}  & \textbf{KLD$\downarrow$}   & \textbf{ACC$\downarrow$}  & \textbf{Time$\downarrow$}  & \textbf{PPL$\downarrow$}   & \textbf{KLD$\downarrow$}   & \textbf{ACC$\downarrow$}  & \textbf{Time$\downarrow$}     \\ \hline
\textbf{Basic}    & \cellcolor{softred}24.78 & \cellcolor{softred}0.975 & \cellcolor{softred}0.670 & \cellcolor{pastelgreen}1.65 & \cellcolor{softred}54.64 & \cellcolor{softred}1.044 & 0.760 & \cellcolor{pastelgreen}1.56 \\
\textbf{MWIS}     & 12.17 & 0.185 & 0.625 & 4.28 & 27.00 & 0.175 & \cellcolor{softred}0.762 & \cellcolor{softred}7.87 \\
\textbf{SyncPool} & 17.55 & \cellcolor{pastelgreen}0.105 & 0.615 & \cellcolor{softred}4.73 & 41.03 & \cellcolor{pastelgreen}0.094 & 0.755 & 4.98 \\
\textbf{Stepwise verification}     & \cellcolor{pastelgreen}12.07 & 0.152 & \cellcolor{pastelgreen}0.597 & 4.15 & \cellcolor{pastelgreen}24.76 & 0.111 & \cellcolor{pastelgreen}0.695 & 5.18 \\
\midrule[1.0pt]
         & \multicolumn{4}{c|}{\textbf{4.0 $\leq$ BPT $<$ 5.0}} & \multicolumn{4}{c}{\textbf{5.0 $\leq$ BPT $<$ 6.0}}  \\ \cline{2-9}
         & \textbf{PPL$\downarrow$}  & \textbf{KLD$\downarrow$}   & \textbf{ACC$\downarrow$}  & \textbf{Time$\downarrow$}  & \textbf{PPL$\downarrow$}   & \textbf{KLD$\downarrow$}   & \textbf{ACC$\downarrow$}  & \textbf{Time$\downarrow$}  \\ \hline
\textbf{Basic}    & \cellcolor{softred}113.02 & \cellcolor{softred}1.122 & \cellcolor{softred}0.796 & \cellcolor{pastelgreen}2.69 & \cellcolor{softred}232.97 & \cellcolor{softred}1.216 & \cellcolor{softred}0.806 & \cellcolor{pastelgreen}5.03 \\
\textbf{MWIS}     & 53.70 & 0.188 & 0.758 & \cellcolor{softred}9.75 & 111.11 & 0.193 & 0.686 & \cellcolor{softred}13.93 \\
\textbf{SyncPool} & 85.36 & \cellcolor{pastelgreen}0.088 & 0.750 & 6.53 & $-$ & $-$ & $-$ & $-$ \\
\textbf{Stepwise verification}     & \cellcolor{pastelgreen}50.08 & 0.093 & \cellcolor{pastelgreen}0.699 & 6.68 & \cellcolor{pastelgreen}102.78 & \cellcolor{pastelgreen}0.075 & \cellcolor{pastelgreen}0.667 & 8.25 \\
\bottomrule[1.0pt]
\end{tabular}}
 \caption{Quantitative comparison with previous disambiguation methods on Qwen2.5-7b and Chinese contexts.}
 \label{table: steganography_Qwen_results}
\end{table*}

\section{Experiments}
In this section we explore the behaviors of addressing TI in steganography and watermarking, using Llama-2-7b~\cite{touvron2023llama} (with English contexts), Swallow-7b~\cite{Fujii:COLM2024,Okazaki:COLM2024} (with Japanese contexts) and 
Qwen2.5-7b~\cite{qwen2.5,qwen2} (with Chinese contexts) (the same as Section~\ref{sec: investigation}). The prompts are randomly selected from the multilingual C4 dataset~\cite{2020t5}. Implementation details can be found in Appendix~\ref{sec:overall_setup}.

\subsection{Experiments on Steganography}
The secret message for embedding is a random 128-bit message, i.e. $m \sim$ Uniform$(\{0,1\}^{128})$. 
As our proposed stepwise verification method and those disambiguation methods (Section~\ref{sec: Disambiguation Algorithms}) all enable steganographic extraction 100\% correct, it is reasonable to use these 100\% disambiguation algorithms as baseline methods, namely, Basic~\cite{nozaki-murawaki-2022-addressing}, MWIS~\cite{yan2023A}, and SyncPool~\cite{qi2024provably}. 

All methods employ arithmetic coding~\cite{ziegler-etal-2019-neural},
an efficient attempt to provably secure steganography~\cite{10179287}.
To evaluate performance under varying embedding capacities, experiments are conducted with different top-$k$ sampling values ($k$ $\in \{$4, 8, 16, 32, 64, 128, 256, 512, 1024, 2048, 4096$\}$). For each top-$k$ value, for each disambiguation method and ours, and for each language model, 500 samples are generated.

\subsubsection{Metrics}
Bits per token (\textbf{BPT}) is a fundamental metric in linguistic steganography, measuring the embedding capacity.
% Steganographic extraction error rate (\textbf{ER}) indicates the rate that incorrect extraction occurs.
Perplexity (\textbf{PPL}) assesses the quality and fluency of the generated text. 
KL divergence (\textbf{KLD}) between modified and original candidate pools quantifies statistical disparities, reflecting imperceptibility. 
Steganalysis accuracy (\textbf{ACC}) indicates the ability to invalidate detection, which is evaluated using a fine-tuned discriminator, with further details provided in Appendix~\ref{sec: steganalysis}. 
Finally, the running time (\textbf{Time}, in seconds) to embed a secret message indicates the steganographic efficiency.

\subsubsection{Results}
While only 128 bits are embedded, there are non-negligible extraction error rates for all the three adopted language models, which are about 10\% for Llama-2-7b, about 5\% for Swallow-7b, and about 7\% for Qwen2.5-7b (details are shown in Appendix~\ref{sec: original error rates}). This illustrates the necessity of our stepwise verification method or disambiguation methods for steganography.

For each method, average experimental data obtained under various top-$k$ values are grouped into embedding-capacity intervals (2.0 $\leq$ \textbf{BPT} $<$ 6.0). Tables~\ref{table: steganography_llama_results},~\ref{table: steganography_swallow_results}, and~\ref{table: steganography_Qwen_results} show the average performance across these intervals for each approach. Note that when the sample size in any group is 20 or fewer, the data is considered insufficient and marked as ``$-$'' in these tables. For each group, within these valid data, the best data point is marked in \colorbox{pastelgreen}{green background} and the worst data point is marked in \colorbox{softred}{red background}. The main findings via these experiments are as follows:

\textit{1)} One of the baseline methods, SyncPool~\cite{qi2024provably}, severely suffers from limitations in embedding capacity. Especially in Llama-2-7b and Swallow-7b, only when \textbf{BPT} $<$ 3.0, there are sufficient data. Details and explanations of this problem are shown in Appendix~\ref{sec: SyncPool limitation}.

\textit{2)} The operation efficiency of our stepwise verification method is acceptable. In our method, it is necessary to call the tokenizer to check each candidate token. Compared to baseline methods whose time complexity is at least $\mathrm{O}(n^2)$ ($n$ is $k$ in the case), the lightweight aspect of our complexity is $\mathrm{O}(n)$. According to the \textbf{Time} dimension, our method is much more efficient than MWIS~\cite{yan2023A}. Even though the Basic~\cite{nozaki-murawaki-2022-addressing} method is the most efficient, the gap between it and ours could be narrowed as \textbf{BPT} increases essentially as $k$ increases. How \textbf{BPT} varies as $k$ varies can be found in Figure~\ref{fig: plot_topk} in the Appendix.

\textit{3)} Overall, our method outperforms the baselines. Compared to the best baseline method for each metric for each language model in each interval, ours achieves an average reduction of 14.12\% in \textbf{PPL}, 47.86\% in \textbf{KLD}, and 3.53\% in \textbf{ACC}.

\begin{table*}[!t]
\small
\renewcommand{\arraystretch}{1.0}
\centering

  \scalebox{1.0}{
\begin{tabular}{l|l|rrr|rr|rr}
\toprule[1.0pt]
\multicolumn{2}{c|}{\multirow{2}{*}{Watermarking scheme}} & \multicolumn{3}{c|}{\textbf{Unattacked}}                                             & \multicolumn{2}{c|}{\textbf{Attacked} ($\epsilon = 0.2$)}  & \multicolumn{2}{c}{\textbf{Attacked} ($\epsilon = 0.4$)}                                        \\ \cline{3-9} 
\multicolumn{2}{c|}{}                                     & \begin{tabular}[c]{@{}r@{}}Watermark\\ Strength$\uparrow$\end{tabular} & AUROC$\uparrow$ & PPL$\downarrow$ & \begin{tabular}[c]{@{}r@{}}Watermark\\ Strength$\uparrow$\end{tabular} & AUROC$\uparrow$ & \begin{tabular}[c]{@{}r@{}}Watermark\\ Strength$\uparrow$\end{tabular} & AUROC$\uparrow$ \\ \midrule[1.0pt]
\multirow{2}{*}{\textbf{LeftHash}} & Original       & 7.58  &  0.996                                                        & 20.55 & 4.59  & 0.982 & 2.57 & 0.878  \\
                          & Post-hoc rollback & \textbf{7.73}    &   0.996                                                     & \textbf{19.56} & \textbf{4.82}  & \textbf{0.984}  & \textbf{2.58} & \textbf{0.879}  \\ \hline
\multirow{2}{*}{\textbf{SelfHash}} & Original       & 7.33   &  0.999                                                       & 20.53 & 4.09  & 0.967 & 2.01  & 0.812   \\
                          & Post-hoc rollback & \textbf{7.44}  &  0.999                                                        & \textbf{19.87} & \textbf{4.30}   & \textbf{0.973}  & \textbf{2.05}  & \textbf{0.820}  \\ \hline
\multirow{2}{*}{\textbf{Unigram}}  & Original       & 7.76  & \textbf{0.998}                                                         & 19.00 & 6.43  & 0.987 & 5.18  & \textbf{0.912}   \\
                          & Post-hoc rollback & \textbf{7.77}  & 0.995                                                        & \textbf{17.85}  & \textbf{6.50}  & 0.987 & \textbf{5.23}  & 0.910  \\ \hline
\multirow{2}{*}{\textbf{Gumbel}}   & Original       & 21.43 &  0.952                                                         & 3.47 & 13.93 & 0.918 & 8.64 & 0.843  \\
                          & Post-hoc rollback & \textbf{23.60} &  \textbf{0.956}                                                         & \textbf{3.15} & \textbf{15.29} & \textbf{0.935} & \textbf{9.62} & \textbf{0.880}  \\
\bottomrule[1,0pt]
\end{tabular}}
 \caption{Quantitative comparison in various watermarking schemes on Llama-2-7b and English context.}
 \label{table: watermark_llama_results}
\end{table*}

\subsection{Experiments on Watermarking}
We implement our proposed post-hoc rollback method for two types of LLM watermarking, which are, respectively, (1) logit-based watermarking and (2) sampling-based watermarking. 

For logit-based watermarking, the LeftHash scheme~\cite{pmlr-v202-kirchenbauer23a} (the context width is 1), the SelfHash scheme~\cite{kirchenbauer2024on} (the context width is 4), and the Unigram scheme~\cite{zhao2024provable} are adopted. For each method, we set green list size $\gamma = 0.5$, and hardness parameter (priority in logits of green list) $\delta = 2.0$.
For sampling-based watermarking, the Gumbel softmax scheme~\cite{Aaronson}\footnote{We use the version  provided by \citet{fu-etal-2024-gumbelsoft}.} is adopted (the context width is 5). Detailed schemes are shown in Appendix~\ref{sec: watermarking schemes}.

Due to differences in temporary inconsistency rates between various language models (shown in Table~\ref{table:temporary_glitch_rate}), and this rate in Llama-2-7b is much lower than those of the other two models. The observation period $q$ is also set differently according to these models, i.e. larger $q$ representing higher relaxation, is set for scenarios with higher temporariness. 
Specifically, $q = 2$ is set for Llama-2-7b, and $q = 10$ is set for Swallow-7b and Qwen2.5-7b.\footnote{How to determine $q$ is detailed in Appendix~\ref{sec: determine q}.}
Besides, the number of generated tokens is constantly 200.
% For whether to use our post-hoc rollback method or not, for each watermarking method, and for each language model, 500 samples are generated and collected.
Regardless of whether to use our post-hoc rollback method, 500 samples are generated and collected for each watermarking method and each language model.

\subsubsection{Attacking Watermarks}
Considering the scales, fairness, and availability of attacking models, we adopt the original language model that generates watermarked texts as the replacement model to attack watermarks.\footnote{For watermarking attacks, the common T5 model~\cite{2020t5} does not support Japanese or Chinese.} 
For each token, it can be selected and then replaced by inference (according to the left context) and resampling, where the selection probability is $\epsilon$.  Besides, for more practical scenarios, results under GPT-4o paraphrasing attack are shown in Appendix~\ref{sec: GPT-4o Attack}.

\subsubsection{Metrics}
% Consider a watermarked text $[s^{(1)},...,s^{(T)}]$, its watermark strength (an indicator of detectability) is denoted as $\Phi(s^{(1)},...,s^{(T)})$.
% For logit-based watermarking, the watermark strength for detection is $z$-score. Specifically,
% \begin{equation}\label{eq: logit_score}
% \Phi(s^{(1)},...,s^{(T)}) = z =\frac{|s|_G - \gamma T}{\sqrt{T\gamma(1-\gamma)}}
% \end{equation}
% where $|s|_G$ is the number of green list tokens, and $T$ is the number of total tokens.

% For sampling-based watermarking, the watermark strength for detection is:
% \begin{equation}\label{eq: sample_score}
% \Phi(s^{(1)},...,s^{(T)}) = \frac{1}{\sqrt{T}} \sum_{t = 1}^{T} \phi^{(t)} - \sqrt{T}
% \end{equation}
% where $s_t$ reflects the correlation between the watermark key and the token $\phi^{(t)}$, and more details are shown in Appendix~\ref{sec: watermarking schemes}.

The detectability of watermarked texts is denoted by \textbf{watermark strength}, where a higher watermark strength increases the likelihood of the text being detected as watermarked. Details about how to calculate the strength are shown in Appendix~\ref{sec: watermarking schemes}.
Due to differences in approaches to computing watermark strength, the obtained watermark strengths indicate relative scores and are not comparable across different watermarking types. Besides, \textbf{AUROC} value is employed to simulate detectability in real-world scenarios, where 500 watermarked texts and 500 unwatermarked texts are evaluated in each group.
Perplexity (\textbf{PPL}) assesses the text quality. 

\subsubsection{Results}
Table~\ref{table: watermark_llama_results} lists the average experimental statistics in various watermarking methods under Llama-2-7b.\footnote{For Swallow-7b and Qwen2.5-7b, the results are shown in Tables~\ref{table: watermark_swallow_results} and~\ref{table: watermark_qwen_results} (in the Appendix).} The main findings are:

\textit{1)} The watermark strengths with the rollback mechanism exhibit a steady increase compared to the original. This aligns with the fact that inconsistent tokens interfere with the detection process. However, the extend of improvement is limited due to the infrequency of inconsistent tokens.
% (discussed in Section~\ref{sec: investigation} with Tables~\ref{table:token_level_rate} and~\ref{table:candidate_level_rate}). 

\textit{2)} Some superiority of our post-hoc rollback method in watermark strengths and AUROC values under attack scenarios represents that watermarks after addressing TI are more detectable and more robust, i.e. higher anti-modification capability.

\textit{3)} \textbf{PPL} in each group counterintuitively decreases steadily, as our method does not target it.\footnote{Perplexities in attack scenarios are shown in Appendix~\ref{sec: Perplexities of Attacked Watermarked Texts}.} One possible explanation for this could be: The tokenizer-based perplexity calculation is affected by inconsistent tokens. Specifically, the predicted probabilities of CITs could be considered very low, thus \textbf{PPL} becomes higher finally.\footnote{How inconsistent tokens affect perplexity is detailed in Appendix~\ref{sec: samples of effects on perplexities}.}

\section{Conclusion}
We observed that tokenization inconsistency (TI) in LLM-based steganography and watermarking can cause robustness issues in extraction or detection processes.
Our investigation on inconsistent tokens across different LLMs and language contexts reveals the infrequency and temporariness of inconsistent tokens.
Based on these two characteristics, we propose two methods to address TI: one for steganography and the other for watermarking.
Our experiments, conducted across various language models, demonstrate that: (1) for steganography, our stepwise verification method outperforms traditional disambiguation approaches across embedding-capacity intervals, offering superior text quality, imperceptibility, and anti-steganalysis capacity; 
(2) for watermarking,  our post-hoc rollback method enhances both detectability and robustness against adversarial modifications while maintaining lower perplexity. 
Our proposed methods have great potential in generalizability, as they can be applied to a wide range of steganographic algorithms and watermarking schemes.
% where texts are generated token by token.

\section*{Acknowledgments}
We express our gratitude to the anonymous reviewers for their valuable and insightful comments.
This work was supported by JST SPRING, Grant Number JPMJSP2110.

\section*{Limitations}

% \textbf{For investigation:} We reported the significant differences between the temporariness of inconsistent tokens in Llama-2-7b and the other two language models. However, we only consider the differences in experimental setups, and do not provide sound explanations for such differences (except for the states about `<s>', `</s>' and partial UTF-8 tokens).

\textbf{For steganography:} 
% The stepwise verification method for steganography is relatively time-consuming compared to disambiguation approaches when the $k$ value of top-$k$ sampling is small, due to the frequent call of tokenizers.
% Even though the stepwise verification method empirically achieves 100\% correct steganographic extraction, there remains an extremely low-probability scenario where extraction errors may occur: 
% If every token in the vocabulary at any generation step is a candidate-level IT, the extraction process must terminate, leading to an extraction failure.
Similar to mainstream linguistic steganographic approaches, the threat model assumes the absence of an active attacker capable of modifying the stegotexts. Otherwise, the guarantee of 100\% correct steganographic extraction would not be ensured.

\textbf{For watermarking:}
Compared to original TI-unaware watermarking schemes, the superiority of the watermarking schemes equipped with the post-hoc rollback method is limited. 
It is the infrequency of inconsistent tokens that makes improvements to various metrics relatively minor.
Specifically, for comparison, any inconsistent token in steganography can disturb subsequent inference.\footnote{The incorrectness of steganographic extraction should refer to text-level inconsistency rates shown in Table~\ref{table:text_level_rate}, which could be higher than 20\%.} The inconsistent tokens in watermarking can merely influence scores at inconsistency positions and positions whose watermarking contexts contain inconsistent tokens during watermarking detection process.\footnote{The token-level inconsistency rates are often below 0.5\% shown in Table~\ref{table:token_level_rate}.}

In this work, the two proposed methods are grounded in an empirical investigation of inconsistent tokens. However, a rigorous theoretical account of how such tokens arise is still lacking, which presents a promising direction for future research.

\section*{Ethical Considerations}
Intended applications of steganography are embedding copyright information, countering censorship, and similar uses. 
However, it can also be used to be exploited for harmful purposes, such as covert communication by malicious actors, spreading disinformation, or bypassing censorship mechanisms. Hence, its potential to facilitate illicit activities necessitates robust monitoring and regulation to prevent misuse.
In addition, countermeasures against steganography, steganalysis, the study of detecting the presence of hidden messages, would also be an encouraging research direction to safeguard against malicious use.

\bibliography{acl_latex}

\newpage

\appendix

\section{Preliminaries \& Related Work}

\subsection{Language Model Basics}

A language model (LM) has a vocabulary $\mathcal{V}$ containing words or word fragments known as ``tokens.''
% The size of typical vocabularies ($|\mathcal{V}|$) is greater than $50{,}000$ tokens~\cite{radford2019language}.
Consider a sequence of LM-generated $T$ tokens $\{s^{(t)}\} \in \mathcal{V}^T$. 
Entries with negative indices, $[s^{(-N_p)},\dots,s^{(-1)}]$, represent a ``prompt'' of length $N_p$ and $[s^{(0)},\dots,s^{(T)}]$ are tokens generated by an LM in response to the prompt.

An LM for the next token prediction at position $t$, is a function $f_{\mathrm{LM}}(\cdot)$ whose input is a sequence of known tokens $[s^{(-N_p)},\dots,s^{(t-1)}]$ which consists of a prompt and the first $t-1$ LM-generated tokens. Then it outputs a logit vector, corresponding to each token 
% $t_i$ (where $i=1,\dots,|\mathcal{V}|$) 
in $\mathcal{V}$.
These logits are then converted into a discrete probability distribution $\boldsymbol{p}^{(t)} = (p^{(t)}_1,\dots,p^{(t)}_{|\mathcal{V}|})$ over the vocabulary, by a $\mathrm{softmax}$ operator (for example).
The next token is then sampled from $\boldsymbol{p}^{(t)}$ using either standard multinomial sampling, beam search, or greedy sampling and so on.

\subsection{Steganography based on Language Models}

% \subsubsection{Overview}
% We give an overview of LM steganography (also a sort of linguistic steganography): 
Alice (the sender) wants to communicate a secret message $m_s \sim$ $U(\{0,1\}^L)$ with Bob (the receiver) by embedding it in a natural language cover text $t_s$ (a stegotext).
The uniform distribution is chosen for $m_s$ without loss of generality: if $m_s$ has additional structure it can be further compressed to a uniformly distributed random variable~\cite{10.1109/TIT.2004.840860}. 
Alice and Bob have agreed on an embedding function $\mathcal{S}_{emb}$ and an extracting function $\mathcal{S}_{ext}$ that perform steganography. Alice and Bob also have access to the exact same language model, $\mathcal{M}^o$, which can be used during embedding and extraction. These two functions are supposed to be invertible. In other words, $\mathcal{S}_{emb}(\mathcal{M}^o,m_s) = t_s$, $\mathcal{S}_{ext}(\mathcal{M}^o,t_s)=m_s'$, and $m_s'$ should be equal to $m_s$.

% \subsubsection{Generative Linguistic Steganography}

Generative linguistic steganography utilizes redundancy of candidate pools to achieve steganography. Through further sampling (e.g. top-$k$) and encoding $\boldsymbol{p}^{(t)}$ with Huffman coding~\cite{8470163} or arithmetic coding~\cite{ziegler-etal-2019-neural} and so on, a steganographic candidate pool $\boldsymbol{\hat{c}}^{(t)}$ is obtained, with its probability distribution $\boldsymbol{\hat{p}}^{(t)}$.
During embedding process, the language model in turn chooses a token in $\boldsymbol{\hat{c}}^{(t)}$ ($t = 0,1, ...$) until it encodes the whole secret message $m_s$;
during extraction process, the language model in turn chooses and extracts a token in $\boldsymbol{\hat{c}}^{(t)}$ ($t = 0,1, ...$)  till the end of the stegotext.

\subsubsection{Segmentation Ambiguity}

The stegotext generated by $\mathcal{S}_{emb}$ is essentially a sequence composed of tokens.
The sender must detokenize it using a tokenizer into a stegotext before transmission.
As shown in Figure~\ref{fig: glitch_token_steganography} (which is also an illustration of segmentation ambiguity), if the sender
generates a token mapping to ``\_no'' and ``body'', the sender needs to detokenize them into the text ``nobody'' before sending it to Bob. However, the issue is that common words like ``\_nobody'' often exist as independent tokens ``\_no'' in the model's vocabulary as well. As a result, a single piece of text can correspond to two or even more different token representations. 
Therefore, during extraction $\mathcal{S}_{ext}(\mathcal{M}^o,t_s)$, since both ``\_nobody'' and ``\_no'' exist in the candidate pool, Bob cannot determine which token the sender embedded the message into.
This phenomenon is referred to as \textit{segmentation ambiguity}. This issue can be exempted in only a few tokenizer-free linguistic steganographic approaches~\cite{math8091558, 10831652}.

\subsubsection{Disambiguation Algorithms}
\label{sec: Disambiguation Algorithms}

Recently, several solutions have emerged to address segmentation ambiguity which achieves 100\% disambiguation in extraction.

\textit{1) Basic Solution:}~\citet{nozaki-murawaki-2022-addressing} proposed a simple disambiguation approach, which removes tokens whose mapping subwords are prefixes of others during every generation and extraction step. 
This process ensures that any token sent by the sender is uniquely extractable for the receiver.

\textit{2) MWIS-based Solution:}~\citet{yan2023A} considered the influence of removing candidate words on the probability distributions and decided to process only if candidate-level ambiguity occurred. Their solution identifies the maximum weight independent set (MWIS) in the candidate pool to reduce probability distortion.

\textit{3) SyncPool Solution:}~\citet{qi2024provably} designed provably secure disambiguation linguistic steganography based on ambiguity pool grouping and synchronous sampling to address information loss and token synchronization issues during steganography, eliminating segmentation ambiguity without altering the distribution.

All of these previous disambiguation approaches make the steganographic extraction merely based on prefixes of stegotexts instead of the tokenizer, bypassing the TI between the sender-receiver pair.\footnote{Another disambiguation method~\cite{10831370} is not introduced in this section or adopted as the baseline in experiments, as its disambiguation is reported to be not 100\%.}
As a result, these traditional disambiguation methods overly process candidate pools, compromising imperceptibility or embedding capacity.

\subsubsection{Imperceptibility of LM-based Steganography}
\label{sec: imperceptibility_GLS}
Following the previous formulation~\cite{dai-cai-2019-towards, shen-etal-2020-near}, statistical imperceptibility refers to the similarity between the true language model $\mathcal{M}^t$ in the monitored channel and $\mathcal{M}^s$ which is the language model $\mathcal{M}^o$ integrated with steganographic algorithms. Specifically, the total variation distance (TVD) is used to measure statistical imperceptibility. Consider the TVD between $\mathcal{M}^t$ and $\mathcal{M}^s$, i.e. $d(\mathcal{M}^t, \mathcal{M}^s)$, by triangle inequality:
\begin{equation}\label{eq: triangle}
    d(\mathcal{M}^t, \mathcal{M}^s) \leq d(\mathcal{M}^t, \mathcal{M}^o) + d(\mathcal{M}^o, \mathcal{M}^s)
\end{equation}
As $d(\mathcal{M}^t, \mathcal{M}^o)$ is a criterion to measure the original language model, which is limited by the research on language models. Thus, $d(\mathcal{M}^o, \mathcal{M}^s)$ is the main focus of linguistic steganography.

According to Pinsker’s inequality~\cite{1201071} and additivity of KL divergence, $d(\mathcal{M}^o, \mathcal{M}^s)$ can be further decomposed in each step, that is:\footnote{Some derivation is omitted here, as details are verified in~\cite{dai-cai-2019-towards, shen-etal-2020-near, 1201071}.}
\begin{equation}\label{eq: KLD}
    d(\mathcal{M}^o, \mathcal{M}^s) \leq \sqrt{\frac{\ln{2}}{2}\sum_{t=1}^{\infty}D_{KL}(\boldsymbol{p}^{(t)}||\boldsymbol{\hat{p}}^{(t)})}
\end{equation}
where $\boldsymbol{p}^{(t)}$ is the original probability distribution at $t^{th}$ step, and $\boldsymbol{\hat{p}}^{(t)}$ is transformed from $\boldsymbol{p}^{(t)}$ via sampling and encoding.
Hence,  linguistic steganography could aim to minimize $D_{KL}(\boldsymbol{p}^{(t)}||\boldsymbol{\hat{p}}^{(t)})$, in order to obtain relative near-imperceptibility.

\subsection{Watermarks for Language Models}

% \subsubsection{Overview}
A watermarking algorithm for language models typically comprises two components: a watermark embedding function $\mathcal{W}_{emb}$ and a watermark detecting function $\mathcal{W}_{det}$~\cite{10.1145/3691626}. $\mathcal{W}_{emb}$ takes a language model $\mathcal{M}^o$ and a watermark message $m_w$ as input and outputs a watermarked text $t_w$, expressed as $\mathcal{W}_{emb}(\mathcal{M}^o, m_w) = t_w$. For the detecting function $\mathcal{W}_{det}$, its input is any text $t$, and its output is its predicted watermark message for the text, denoted $\mathcal{W}_{det} = m_w'$. The watermark message $m_w$ or $m_w'$ can be a Boolean value (True or False) for zero-bit watermarks to indicate whether the text is generated by AI~\cite{pmlr-v202-kirchenbauer23a, kirchenbauer2024on}, and can also be a bit stream for multi-bit watermark usage~\cite{yoo-etal-2024-advancing}.
% \subsubsection{Inference Time Watermarking}
So far, there are two main types of inference-time watermarking algorithms: (1) logit-based watermarking and (2) sampling-based watermarking. 

For the former, those methods refer to inserting $m_w$ into the logit of each generative step by language models~\cite{pmlr-v202-kirchenbauer23a, 10374576,kirchenbauer2024on,lu2024entropy,zhao2024provable}. The trade-off between text quality and detectability should be considered in these watermarks.

For the latter, they do not alter the logits, but utilize the watermark message to guide the sampling process~\cite{Aaronson,pmlr-v247-christ24a,kuditipudi2024robust}. For token-by-token sampling watermarking, the principle of incorporating watermarks during the token-sampling phase is derived from the randomness inherent in token sampling. In this scenario, watermarks can be introduced using a fixed seed, where a pseudo-random number generator produces a sequence of pseudo-random numbers to guide the sampling of each token. For watermark detection, it is only necessary to assess the alignment between the text tokens and the pseudo-random numbers, specifically evaluating whether the choice of each token in the text matches the corresponding value in the random number sequence.

\subsection{Related Work on Abnormal Tokens}
Tokenization stands as a cornerstone in natural language processing, which transforms a continuous text sequence into a list of discrete values called tokens~\cite{wang2024tokenizationmattersdegradinglarge}.

\subsubsection{Glitch Tokens}

Glitch tokens refer to a class of anomalous tokens in LLMs that can trigger unexpected and often erroneous behaviors when processed by LLMs. This issue arises from improper tokenization of raw texts, which can stem from irregularities in the training process, such as underrepresentation in training data or inconsistencies in tokenization~\cite{geiping2024coercing}.

Glitch token and according glitchy phenomena in LLMs are first investigated comprehensively and systematically by~\citet{10.1145/3660799}, where glitch-token symptoms and glitch-token taxonomy are explored, and an efficient glitch-token detection method named GlitchHunter is proposed. GlitchHunter iteratively constructs a token embedding graph and generates candidate glitch token clusters for subsequent detection.

A more advanced and effective detection and mitigation of glitch tokens is proposed and named GlitchProber~\cite{10.1145/3691620.3695060}. This work first reveals the characteristic features induced by glitch tokens on LLMs, which are evidenced by significant deviations in the distributions of attention patterns and dynamic information from intermediate model layers. GlitchProber utilizes small-scale sampling, principal component analysis for accelerated feature extraction, and a simple classifier for efficient vocabulary screening.

Another advancing glitch-token detection method is named GlitchMiner~\cite{wu2024mining}, which is a gradient-based discrete optimization framework that efficiently identifies glitch tokens by introducing entropy as a measure of prediction uncertainty and employing a local search strategy to explore the token space.

% As these existing detection methods are not targeted at designated tasks, specific task-related characteristics they can leverage are limited. Therefore, they tend to generally mitigate glitch tokens and glitchy phenomena instead of addressing glitch tokens.
\subsubsection{Unreachable Tokens}
Besides, `unreachable tokens' are termed by~\citet{land2024fishing}, referring to those tokens that are never produced as a result of tokenizing text. In that work, they test this by checking if decoding a token to a string, and re-tokenizing this string, results in the original token ID. Although they also apply the detokenization-retokenization pipeline, they merely consider that one tested token without contexts.

\subsubsection{Tokenization Inconsistency (TI)}
The importance of tokenization consistency is reported in extractive NLP tasks~\cite{sun-etal-2023-tokenization}. They study the issue of tokenization inconsistency that is commonly neglected in training these models, and reveal that this issue damages the extractive nature of these tasks after the input and output are tokenized inconsistently by the tokenizer, thus leading to performance drop as well as hallucination.

Besides, a recent work~\cite{wang2024tokenizationmattersdegradinglarge} constructs an adversarial dataset, named as \textbf{ADT} (Adversarial Dataset for Tokenizer), which draws upon the vocabularies of various open-source LLMs to challenge LLMs’ tokenization.
That study is the first to investigating LLMs’ vulnerability in terms of challenging their token segmentation, which will shed light on the subsequent research of improving LLMs’ capabilities through optimizing their tokenization process and algorithms.

Correct or consistent tokenization is often overlooked in most tasks. However, in text-based transmission systems (including steganography and watermarking) where texts are transmitted from Alice to Bob, tokenization consistency becomes crucial, as precise transmission is essential for maintaining the integrity of the information.

\begin{table}[!t]
\small
\centering
\begin{tabular}{l|r|r|r|r|r}
\toprule[1pt]
\multicolumn{6}{c}{\textbf{Absence of SITs}} \\ \hline
\textbf{Generated token ids} & ... &  18  & 76 & ... & ...  \\ \hline
\textbf{Retokenized token ids} & ... & 18  & \cellcolor{softred} 325  & 76 & ... \\ \midrule[1pt]

\multicolumn{6}{c}{\textbf{Absence of CITs}} \\ \hline
\textbf{Generated token ids} & ... &  1092  & \cellcolor{softred} 8 & 92 & ...  \\ \hline
\textbf{Retokenized token ids} & ... & 1092  &  92  & ... & ...  \\ \bottomrule[1pt]

\end{tabular}
 \caption{Examples of the absence of SITs or CITs when TI occurs.}
 \label{table: proposition 1}

\end{table}

\section{Analysis of Inconsistent Tokens and Tokenization Inconsistency (TI)}
\label{sec: Analysis of Inconsistent Tokens and Tokenization Inconsistency (TI)}
In this section, we provide detailed analysis and explanations of the relationships between source inconsistent tokens (SITs), consequential inconsistent tokens (CITs), candidate-level inconsistent tokens (candidate-level ITs), and tokenization inconsistency (TI).
\begin{proposition}\label{proposition: sufficeint for TI}
The sufficient condition for the existence of TI is the existence of SIT(s) or CIT(s).
\end{proposition}
According to Proposition~\ref{proposition: sufficeint for TI}, when TI occurs, there could be only SITs (in the generated token list) or only CITs (in the retokenized token list). Table~\ref{table: proposition 1} provides examples of TI where SITs or CITs are absent. Inconsistent tokens are marked in \colorbox{softred}{red background}, and other tokens including `...' are all consistent tokens. The essential reason for it is related to tokenizer preferences. For example, the sole SIT or sole CIT could be detokenized to a 0-length character.

\begin{proposition}\label{proposition: for candidate-level ITs and TI}
A necessary condition for the existence of TI is outputting candidate-level IT(s).
\end{proposition}
According to Algorithm~\ref{algorithm: Identify a candidate-level glitch token}, if outputting a candidate token changes the tokenization state from consistency to inconsistency or persists TI, that candidate token is a candidate-level IT. Therefore, an easy proof of Proposition~\ref{proposition: for candidate-level ITs and TI} by contradiction is as follows: 
\textit{If a candidate-level IT is never generated, TI never occurs.}

\begin{proposition}\label{proposition: for candidate-level ITs and SITs}
If all the inconsistent tokens are not temporary, there is still possibility that a candidate-level IT does not become an SIT.
\end{proposition}

According to Proposition~\ref{proposition: for candidate-level ITs and TI}, outputting candidate-level ITs are necessary for TI, and according to Proposition~\ref{proposition: sufficeint for TI}, SITs are not necessary for TI. Therefore, there are some TI cases where candidate-level ITs are output, but SITs are absent.
TI with the absence of SITs (Table~\ref{table: proposition 1}) provides an example of Proposition~\ref{proposition: for candidate-level ITs and SITs}, where the `id: 18' token or the `id: 76' token should an output candidate-level IT, but neither of them is an SIT.

\begin{algorithm}[!t]
\small
\caption{Stepwise verification (embedding)}\label{algorithm: address_glitch_for_steganography_embedding} 
\textbf{Input:}\\
Prompt (initial historical tokens), $[s^{(-N_p)},\dots,s^{(-1)}]$ \\
Secret message, $m_s$  \\
{\textbf{Output:}}\\
Steganographic text, $t_s$\\

\begin{algorithmic}[1]
% \STATE Initialize historical tokens $$ using $s^{(-N_p)},\dots,s^{(-1)}$.
\FOR{$t = 0,1,\dots$}
    \STATE Apply the language model to historical tokens to obtain the probability distribution $\boldsymbol{p}^{(t)}$ over the vocabulary $\mathcal{V}$.
    \STATE Sample $\mathcal{V}$ according to $\boldsymbol{p}^{(t)}$ to get the candidate pool $\boldsymbol{\hat{c}}^{(t)}$.
    \STATE Filter out candidate-level ITs in $\boldsymbol{\hat{c}}^{(t)}$ to get a inconsistency-free candidate pool $\boldsymbol{\hat{c}'}^{(t)}$.
    \IF{$\boldsymbol{\hat{c}'}^{(t)} == \emptyset$}
        \STATE Add the highest probability token (which is not an SIT) from $\mathcal{V}_{\setminus \boldsymbol{\hat{c}}^{(t)}}$ to $\boldsymbol{\hat{c}'}^{(t)}$.
    \ENDIF
    \STATE Get the normalized probability distribution $\boldsymbol{\hat{p}'}^{(t)}$ over $\boldsymbol{\hat{c}'}^{(t)}$.
    \STATE Use the steganographic embedding algorithm and $m_s$ to generate the next token $s^{(t)}$.
    \STATE Append $s^{(t)}$ to the historical tokens.
\ENDFOR

\STATE Detokenize historical tokens to $t_s$.
\RETURN $t_s$

\end{algorithmic}

\end{algorithm}

\begin{algorithm}[!t]
\small
\caption{Stepwise verification (extraction)}\label{algorithm: address_glitch_for_steganography_extraction} 
\textbf{Input:}\\
Prompt (initial historical tokens), $[s^{(-N_p)},\dots,s^{(-1)}]$ \\
Steganographic text, $t_s$\\
{\textbf{Output:}}\\
Secret message, $m_s$  \\

\begin{algorithmic}[1]
% \STATE $\mathcal{H}^0 \leftarrow \mathcal{P}$; \textit{/* Initialize historical tokens*/}
\STATE Tokenize $t_s$ to token list $[s^{(-N_p)},\dots,s^{(0)},s^{(1)},\dots]$.
\FOR{$t = 0,1,\dots$}
    \STATE Apply the language model to historical tokens to obtain the probability distribution $\boldsymbol{p}^{(t)}$ over the vocabulary $\mathcal{V}$.
    \STATE Sample $\mathcal{V}$ according to $\boldsymbol{p}^{(t)}$ to get the candidate pool $\boldsymbol{\hat{c}}^{(t)}$.
    \STATE Filter out candidate-level ITs in $\boldsymbol{\hat{c}}^{(t)}$ to get a inconsistency-free candidate pool $\boldsymbol{\hat{c}'}^{(t)}$.
    \IF{$\boldsymbol{\hat{c}'}^{(t)} == \emptyset$}
        \STATE Add the highest probability token (which is not an SIT) from $\mathcal{V}_{\setminus \boldsymbol{\hat{c}}^{(t)}}$ to $\boldsymbol{\hat{c}'}^{(t)}$.
    \ENDIF
    \STATE Get the normalized probability distribution $\boldsymbol{\hat{p}'}^{(t)}$ over $\boldsymbol{\hat{c}'}^{(t)}$.
    \STATE Use the steganographic extraction algorithm and $s^{(t)}$ to update $m_s$.
\ENDFOR

\RETURN $m_s$

\end{algorithmic}

\end{algorithm}

\begin{algorithm}[!t]
\small
\caption{Post-hoc rollback}\label{algorithm: address_glitch_for_watermark} 
\textbf{Input:}\\
Prompt (initial historical tokens), $[s^{(-N_p)},\dots,s^{(-1)}]$ \\
Watermark message, $m_w$  \\
Observation period parameter, $q$ \\
{\textbf{Output:}}\\
Watermarked text, $t_w$\\

\begin{algorithmic}[1]
\STATE $q_c \leftarrow \mathrm{NULL}$; \\\textit{/* Initialize the state of observation period*/}
% \STATE $\mathcal{H}^0 \leftarrow \mathcal{P}$; \textit{/* Initialize historical tokens*/}
\FOR{$T = 0,1,\dots$}
    \STATE Apply the language model to historical tokens and watermark embedding algorithm to generate the next token $s^{(t)}$;
    \STATE Append $s^{(t)}$ to historical tokens;
    \IF{Tokenization consistency}
        \STATE $q_c \leftarrow \mathrm{NULL}$;
    \ELSE
        \IF{$q_c == \mathrm{NULL}$}
            \STATE $q_c \leftarrow 0$;
        \ENDIF
        \IF{$q_c \neq \mathrm{NULL}$}
            \IF{$q_c < q$}
                \STATE $q_c \leftarrow q_c + 1$;
            \ELSE
                \STATE Delete the latest $(q+1)$ historical tokens;
            \ENDIF
        \ENDIF
    \ENDIF
\ENDFOR

\STATE Detokenize historical tokens to $t_w$;
\RETURN $t_w$

\end{algorithmic}

\end{algorithm}

\section{Algorithms of Methods}

\subsection{Stepwise Verification}
\label{sec: details_stepwise}
Algorithm~\ref{algorithm: address_glitch_for_steganography_embedding} provides details of the steganographic embedding process equipped with our proposed stepwise verification method. This algorithm considers an error scenario with a very small probability of occurrence, that is, the inconsistency-free candidate pool $\boldsymbol{\hat{c}'}^{(t)}$ is $\emptyset$ (Line 4-5). Once it occurs, a non-SIT token outside the steganographic candidate pool $\boldsymbol{{c}'}^{(t)}$ should be added to $\boldsymbol{\hat{c}'}^{(t)}$, to make sure the generation is always able to continue (Line 6).
Algorithm~\ref{algorithm: address_glitch_for_steganography_extraction} provides the details of the extraction version, and also includes the error prevention mechanism (Line 6-7). At each step of generation, both in embedding and extraction, they share the same view of how the candidate pool is processed.

\subsection{Post-Hoc Rollback}
\label{sec: details_posthoc}
Algorithm~\ref{algorithm: address_glitch_for_watermark} provides details on how to implement our proposed post-hoc rollback method in the generation process, meanwhile embedding watermarking. $q_c$ is a signal of whether the generation is currently in the state of TI ($q_c == \mathrm{NULL}$ indicates tokenization consistency). Whenever tokenization consistency is recovered, $q_c$ is reset as $\mathrm{NULL}$ (Line 5-6). Once the tokenization state changes from consistency to inconsistency, $q_c$ is set as $0$ (Line 9), and $q_c$ increases when the inconsistency lasts after generating the next token (Line 12). Once $q_c$ is not less than the designated observation period parameter $q$, the rollback mechanism is triggered: The token-by-token generation rollbacks back to the state where the stable inconsistent token $q$ tokens ago is not generated (Line 14).

\section{Experimental Details}

\subsection{Overall Setups}
\label{sec:overall_setup}

The initial contexts are randomly selected from the multilingual C4 dataset.\footnote{\href{https://huggingface.co/datasets/allenai/c4}{https://huggingface.co/datasets/allenai/c4}} 
The temperature parameter is set to 1.0 constantly.
According to the features of different languages, for Llama-2-7b, the initial 10 words of an item in the C4 dataset are the initial context for each generation; while for Swallow-7b and Qwen2.5-7b, the initial 10 characters of an item in the C4 dataset are the initial context for each generation. 
The perplexity of a text is calculated by the language model that generates the text. 
All the parameters of the tokenizer functions are default, except for setting \texttt{skip\_special\_tokens = True} in detokenization.

All experiments are implemented in Python 3.12.7 with Torch 2.5.0, running on a 2.0 GHz CPU and accelerated by using 8 × NVIDIA RTX A6000 GPUs.

\subsection{Steganalysis}
\label{sec: steganalysis}
As a discriminator for each language, we used a base-sized BERT model taken from Hugging Face's transformers package (English: bert-base-uncased,\footnote{\href{https://huggingface.co/google-bert/bert-base-uncased}{https://huggingface.co/google-bert/bert-base-uncased}} Japanese: cl-tohoku/bert-japanese,\footnote{\href{https://github.com/cl-tohoku/bert-japanese}{https://github.com/cl-tohoku/bert-japanese}} Chinese: bert-base-chinese).\footnote{\href{https://huggingface.co/google-bert/bert-base-chinese}{https://huggingface.co/google-bert/bert-base-chinese}}
Positive samples are collected from stegotexts generated using various top-$k$ samplings, while negative samples are sourced from non-steganographic texts (generated by the same models without any steganographic algorithm). 

As for each top-$k$ sampling value ($k$ $\in \{$4, 8, 16, 32, 64, 128, 256, 512, 1024, 2048, 4096$\}$ - 11 different $k$), for each disambiguation method and ours (4 methods), and for each language model, 500 samples are generated, for each language model the size of collected stegotexts is $11 \times 4 \times 500 = 22000$.
Hence, for each language model, during the training phase, both positive and negative samples consist of 17,600 instances each (80\% of all collected stegotexts). For testing, 4,400 untrained positive samples are used (20\% of all collected stegotexts), categorized into different embedding-capacity intervals as shown in Tables~\ref{table: steganography_llama_results},~\ref{table: steganography_swallow_results}, and~\ref{table: steganography_Qwen_results}. In each embedding-capacity interval and for each disambiguation approach, only stegotexts with a sample size greater than 20 are included in the tests; otherwise, ``$-$'' is marked to indicate insufficient data.

\begin{table}[!t]
\small
\renewcommand{\arraystretch}{1.0}
\centering

  \scalebox{1.0}{
\begin{tabular}{l|rr|rr|rr}
\toprule[1.0pt]
     & \multicolumn{2}{c|}{\textbf{Llama-2-7b}} & \multicolumn{2}{c|}{\textbf{Swallow-7b}} & \multicolumn{2}{c}{\textbf{Qwen2.5-7b}} \\ \midrule[1.0pt]
\textbf{$k$}    & \textbf{BPT}             & \textbf{ER}            & \textbf{BPT}             & \textbf{ER}            & \textbf{BPT}            & \textbf{ER}            \\ \hline
4   & 1.11               & 20.0\%             & 0.95               & 6.0\%             & 1.01              & 14.0\%             \\
8   & 1.62               & 14.6\%             & 1.38               & 5.0\%             & 1.45              & 6.8\%             \\
16   & 2.03               & 11.4\%             & 1.77               & 4.0\%             & 1.98              & 7.4\%             \\
32   & 2.34               & 8.0\%             & 2.12               & 4.4\%             & 2.37              & 6.2\%             \\
64   & 2.64               & 9.4\%             & 2.45               & 3.2\%             & 2.77              & 8.8\%             \\
128   & 2.87               & 8.0\%             & 2.68               & 5.2\%             & 3.08              & 7.2\%             \\
256  & 3.08               & 9.2\%             & 2.94               & 3.6\%             & 3.43              & 7.4\%             \\
512   & 3.21              & 11.4\%             & 3.23               & 4.6\%             & 3.71              & 6.8\%             \\
1024 & 3.27               & 7.2\%             & 3.38               & 4.6\%             & 4.05              & 8.0\%             \\
2048   & 3.41               & 6.2\%             & 3.71               & 5.4\%             & 4.25              & 7.6\%             \\
4096 & 3.41               & 11.4\%             & 3.69               & 3.4\%             & 4.36              & 7.6\%            \\ \bottomrule[1.0pt]
\end{tabular}}
 \caption{Embedding capacities and error rates of steganography (without disambiguation or stepwise verification) implemented under various language models and top-$k$ sampling values.}
 \label{table: original ER}
\end{table}

\begin{figure}[!t]
    \centering
    \begin{subfigure}[b]{\columnwidth}
        \includegraphics[width=\textwidth]{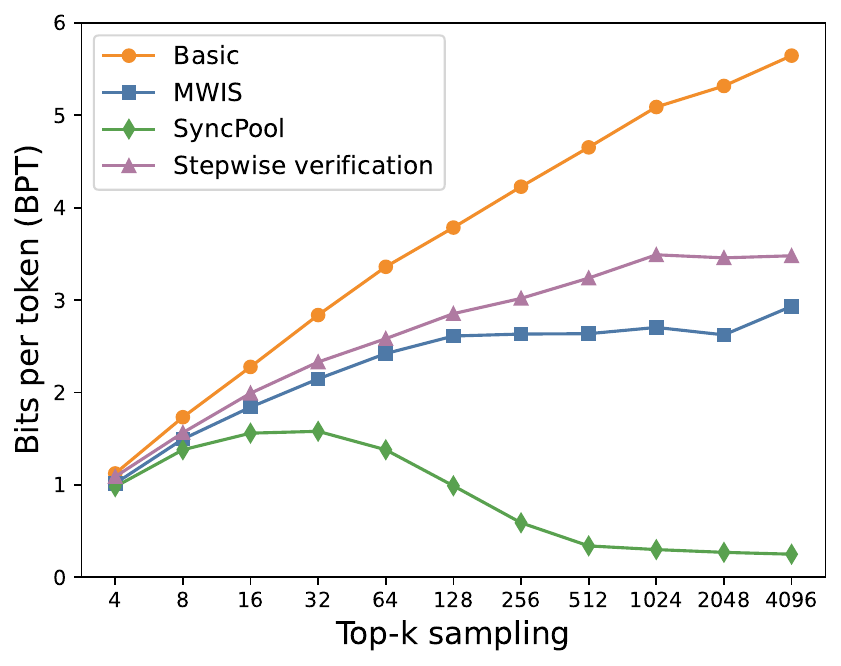}
        \caption{Llama-2-7b.}
        \label{fig: plot_topk_Llama}
    \end{subfigure}
    \hfill
    \begin{subfigure}[b]{\columnwidth}
        \includegraphics[width=\textwidth]{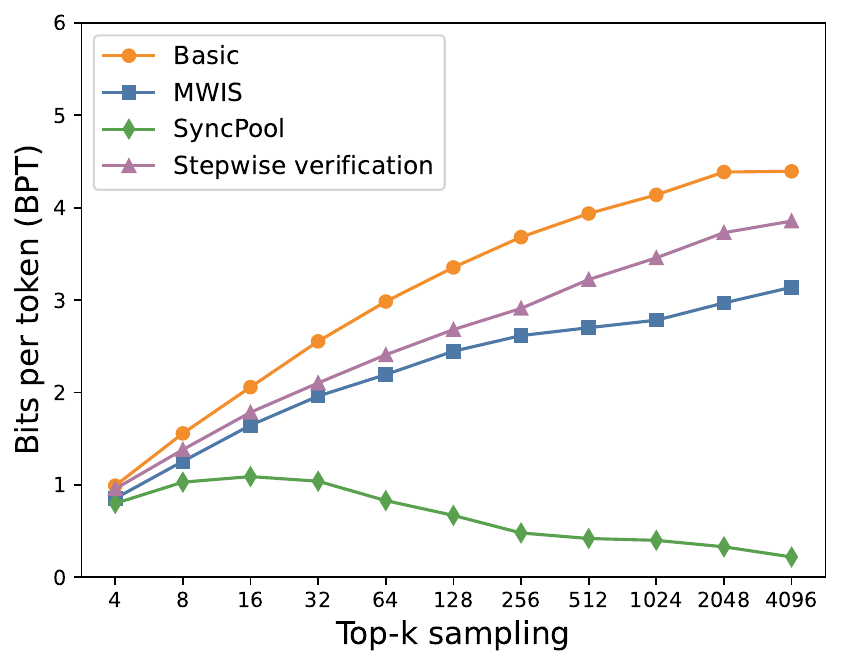}
        \caption{Swallow-7b.}
        \label{fig: plot_topk_Swallow}
    \end{subfigure}
    \hfill
    \begin{subfigure}[b]{\columnwidth}
        \includegraphics[width=\textwidth]{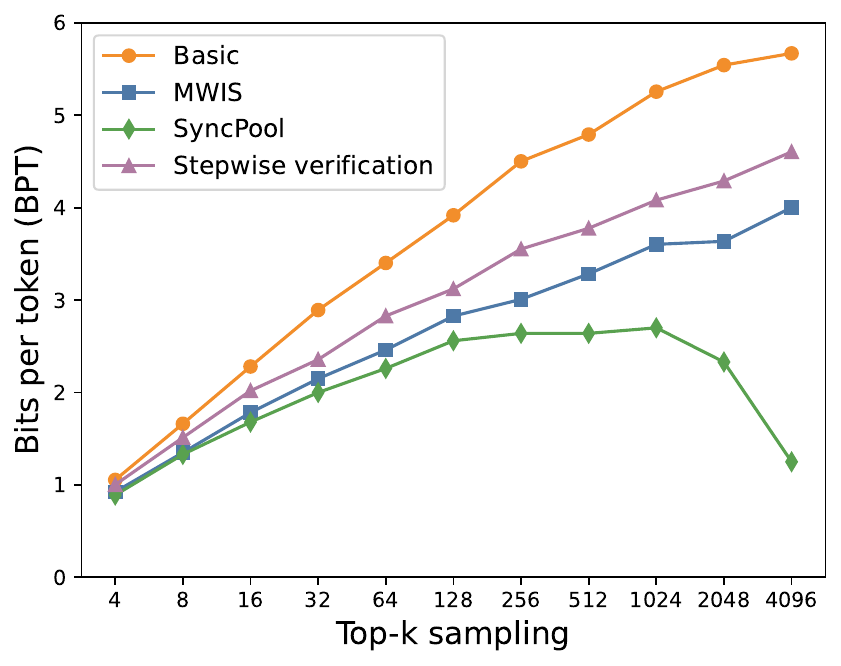}
        \caption{Qwen2.5-7b.}
        \label{fig: plot_topk_Qwen}
    \end{subfigure}
    \caption{Average embedding capacities (bits per token, BPT) when using various disambiguation methods and our stepwise verification method, under top-$k$ values.}
    \label{fig: plot_topk}
\end{figure}

Given the significant variation in the lengths of positive samples, we adjust the negative samples to uniformly vary between 10 and 128 tokens (the prompt is excluded) to ensure that the trained discriminator is not sensitive to text length. Additionally, all texts are padded or truncated to 128 tokens, so that positive samples cannot be distinguished as steganographic based solely on their length.
For fine-tuning the BERT model, we use Adam~\cite{kingma2017adammethodstochasticoptimization} as the optimizer with a learning rate of $5 \times 10^{-5}$. The batch size is set to 2048, and the discriminator is trained for 20 epochs, running time of the whole training process is approximately 10 minutes.

\subsection{Original Error Rates}
\label{sec: original error rates}
In this section, we use empirical statistics to show the extent to which steganography suffers from extraction errors, if extraction errors are neglected in steganographic approaches. We use the steganographic extraction error rate (\textbf{ER}) to indicate the rate that incorrect extraction occurs, and for approaches equipped with neither disambiguation nor our proposed stepwise verification, the error rates are referred to as original error rates. Table~\ref{table: original ER} lists \textbf{BPT} and \textbf{ER} for steganography with extraction errors.
For each language model and for each top-$k$ value, the data size is 500.
For $k=4$, \textbf{BPT} is the lowest and \textbf{ER} is the highest for each language model. The main reason is that when embedding a 128-bit secret message, lower \textbf{ER} means longer generated stegotext. And according to the relationship between the token number and the text-level inconsistency rate shown in Table~\ref{table:text_level_rate}, it is reasonable for longer stegotexts to suffer from higher \textbf{ER}.
In addition, when the length of the secret message increases, it is reasonable to anticipate that the original error rates for various language models increase further.

% \begin{figure}[!t]
%  \centering
%  \includegraphics[width=\columnwidth]{plot_topk_SyncPool.pdf} 
%  \caption{Average embedding capacities (bits per token, BPT) when using the SyncPool disambiguation method, under various language models and top-$k$ values.}
%  \label{fig: plot_topk_SyncPool}
% \end{figure}

\subsection{The Limitation of SyncPool in Embedding Capacity}
\label{sec: SyncPool limitation}
Figure~\ref{fig: plot_topk} illustrates how embedding capacity varies according to various top-$k$ values when different methods are adopted. For all three language models with respectively English, Japanese and Chinese contexts, these data points in SyncPool share a similar trajectory, that is, when top-$k$ value increases, \textbf{BPT} increases when $k$ is small and decreases when $k$ becomes much larger.
For comparison, in other methods, when top-$k$ value increases, \textbf{BPT} increases steadily.

The reason for the phenomenon is that SyncPool merges the original candidate pools into ambiguous pools for subsequent steganographic processing. However, as the ambiguous pools are formed according to prefix relationships in candidate tokens, when the size of original candidate pools increases, the average tokens in each ambiguous pool also increase. As a result, the size of ambiguous pools could rise more rapidly than the size of original candidate pools, so the average number of ambiguous pools could decrease, thus leading to lower embedding capacity.

Furthermore, the work of SyncPool~\cite{qi2024provably} reports a KL divergence of 0 in their experiments, as their reference candidate pools are based on top-$k$ sampled candidates. However, as outlined in Eq.~\ref{eq: KLD}, to accurately compare the divergence between the original language model and the model used for steganography, we compute \textbf{KLD} using the original candidate pools as references in our experiments (Tables~\ref{table: steganography_llama_results},~\ref{table: steganography_swallow_results}, and~\ref{table: steganography_Qwen_results}).

\subsection{Watermarking Schemes}
\label{sec: watermarking schemes}
Consider a text $[s^{(1)},...,s^{(T)}]$, its watermark strength (an indicator of detectability) is denoted as $\Phi(s^{(1)},...,s^{(T)})$.

For logit-based watermarking, including LeftHash~\cite{pmlr-v202-kirchenbauer23a}, SelfHash~\cite{kirchenbauer2024on} and Unigram~\cite{zhao2024provable} adopted by our experiments, their mechanisms are as follows:
\begin{itemize}
    \item \textbf{Context:} For LeftHash and SelfHash, the context is previous $h$ tokens; there is no context for Unigram.
    \item \textbf{Pseudo-random function:} For LeftHash and SelfHash, $F_{sk}(\mathrm{context})$ hashes the context to a seed at each generative step; Unigram adopts a global seed. Then the seed is used to generate a random vector $\mathrm{Vec}^{(t)}$ in $\{0,1\}^{|\mathcal{V}|}$, the vector has $\gamma|\mathcal{V}|$ 1's (representing green tokens) and $(1-\gamma)|\mathcal{V}|$ 0's (representing red tokens).
    \item \textbf{Decoder:} Sample a token $s^{(t)}$ from $\mathrm{softmax}(\mathrm{logit}^{(t)} + \delta * \mathrm{Vec}^{(t)})$.
    \item \textbf{One-token score:} $\phi^{(t)} = \mathrm{Vec}^{(t)}[s^{(t)}]$.
    \item \textbf{Watermark strength:} $\Phi(s^{(1)},...,s^{(T)}) = \frac{\sum_{t = 1}^{T}\phi^{(t)} - \gamma T}{\sqrt{T\gamma(1-\gamma)}}$.
\end{itemize}

For sampling-based watermarking, the detailed scheme of Gumbel softmax scheme~\cite{Aaronson} is:
\begin{itemize}
    \item \textbf{Context:} The previous $h$ tokens.
    \item \textbf{Pseudo-random function:} $F_{sk}(\mathrm{context})$ hashes the context to a seed at each generative step, then uses the seed to generate a random vector $\mathrm{Vec}^{(t)}$ in $(0,1)^{|\mathcal{V}|}$ where each element is uniformly sampled from $(0,1)$.
    \item \textbf{Decoder:} Select a token  $s^{(t)}$  which is $\mathrm{argmax}_{1\leq i \leq |\mathcal{V}|}\frac{\mathrm{log}(\mathrm{Vec}^{(t)}[i])}{\mathrm{softmax}(\mathrm{logit}^{(t)})[i]}$.
    \item \textbf{One-token score:} $\phi^{(t)} = -\mathrm{log}(1 - \mathrm{Vec}^{(t)}[s^{(t)}])$.
    \item \textbf{Watermark strength:} $\Phi(s^{(1)},...,s^{(T)}) = \frac{1}{\sqrt{T}} \sum_{t = 1}^{T} \phi^{(t)} - \sqrt{T}$.
\end{itemize}

\begin{table}[!t]
\small
\renewcommand{\arraystretch}{1.0}
\centering
\begin{tabular}{l|l|r|r}
\toprule[1.0pt]
                          \multicolumn{2}{l|}{Watermarking scheme}      & \multicolumn{1}{c|}{\begin{tabular}[c]{@{}r@{}}\textbf{Attacked} \\ ($\epsilon = 0.2$)\end{tabular}}  & \multicolumn{1}{c}{\begin{tabular}[c]{@{}r@{}}\textbf{Attacked} \\ ($\epsilon = 0.4$)\end{tabular}} \\ \midrule[1.0pt]
\multirow{2}{*}{\textbf{LeftHash}} & Original & 185.25        & 477.71                                                                              \\
                          & Post-hoc rollback & \textbf{162.56}       &    \textbf{417.41}                                                                            \\ \hline
\multirow{2}{*}{\textbf{SelfHash}} & Original & 188.69        &  470.96                                                                             \\
                          & Post-hoc rollback &   \textbf{172.93}      &   \textbf{455.81}                                                                            \\ \hline
\multirow{2}{*}{\textbf{Unigram}}  & Original & 167.16        &   438.53                                                                            \\
                          & Post-hoc rollback & \textbf{161.24}        &  \textbf{424.75}                                                                             \\ \hline
\multirow{2}{*}{\textbf{Gumbel}}   & Original &  19.38       &  46.32                                                                             \\
                          & Post-hoc rollback &  \textbf{18.70}       & \textbf{40.70}        \\ \bottomrule[1.0pt]                                                                     
\end{tabular}
 \caption{Perplexities in various watermarking schemes under attack scenarios when Llama-2-7b is adopted.}
 \label{table: attack_perplexity_llama}

\end{table}

\begin{table}[!t]
\small
\renewcommand{\arraystretch}{1.0}
\centering
\begin{tabular}{l|l|r|r}
\toprule[1.0pt]
                          \multicolumn{2}{l|}{Watermarking scheme}      & \multicolumn{1}{c|}{\begin{tabular}[c]{@{}r@{}}\textbf{Attacked} \\ ($\epsilon = 0.2$)\end{tabular}}  & \multicolumn{1}{c}{\begin{tabular}[c]{@{}r@{}}\textbf{Attacked} \\ ($\epsilon = 0.4$)\end{tabular}} \\ \midrule[1.0pt]
\multirow{2}{*}{\textbf{LeftHash}} & Original & 133.44        &  301.82                                                                             \\
                          & Post-hoc rollback &  \textbf{130.31}       & \textbf{290.67}                                                                              \\ \hline
\multirow{2}{*}{\textbf{SelfHash}} & Original & 143.03        &  \textbf{308.02}                                                                             \\
                          & Post-hoc rollback &  \textbf{142.34}       &   313.01                                                                            \\ \hline
\multirow{2}{*}{\textbf{Unigram}}  & Original &  136.13       &   301.39                                                                            \\
                          & Post-hoc rollback &  \textbf{134.44}       &   \textbf{291.96}                                                                            \\ \hline
\multirow{2}{*}{\textbf{Gumbel}}   & Original &  5.99       & 9.44                                                                             \\
                          & Post-hoc rollback &  \textbf{5.78}       &  \textbf{9.29}       \\ \bottomrule[1.0pt]                                                                     
\end{tabular}
 \caption{Perplexities in various watermarking schemes under attack scenarios when Swallow-7b is adopted.}
 \label{table: attack_perplexity_swallow}

\end{table}

\begin{table}[!t]
\small
\renewcommand{\arraystretch}{1.0}
\centering
\begin{tabular}{l|l|r|r}
\toprule[1.0pt]
                          \multicolumn{2}{l|}{Watermarking scheme}      & \multicolumn{1}{c|}{\begin{tabular}[c]{@{}r@{}}\textbf{Attacked} \\ ($\epsilon = 0.2$)\end{tabular}}  & \multicolumn{1}{c}{\begin{tabular}[c]{@{}r@{}}\textbf{Attacked} \\ ($\epsilon = 0.4$)\end{tabular}} \\ \midrule[1.0pt]
\multirow{2}{*}{\textbf{LeftHash}} & Original &   414.34      & 962.58                                                                              \\
                          & Post-hoc rollback &  \textbf{378.68}       & \textbf{857.41}                                                                              \\ \hline
\multirow{2}{*}{\textbf{SelfHash}} & Original & 398.99        &  890.30                                                                             \\
                          & Post-hoc rollback &  \textbf{355.39}       &  \textbf{786.71}                                                                             \\ \hline
\multirow{2}{*}{\textbf{Unigram}}  & Original &  \textbf{309.83}       & 686.18                                                                              \\
                          & Post-hoc rollback & 322.16        & \textbf{676.22}                                                                               \\ \hline
\multirow{2}{*}{\textbf{Gumbel}}   & Original &  6.96       & 12.76                                                                              \\
                          & Post-hoc rollback & \textbf{6.69}        &  \textbf{11.57}       \\ \bottomrule[1.0pt]                                                                     
\end{tabular}
 \caption{Perplexities in various watermarking schemes under attack scenarios when Qwen2.5-7b is adopted.}
 \label{table: attack_perplexity_qwen}

\end{table}

% \usepackage{multirow}

\subsection{Perplexities of Attacked Watermarked Texts}
\label{sec: Perplexities of Attacked Watermarked Texts}
Tables~\ref{table: attack_perplexity_llama},~\ref{table: attack_perplexity_swallow}, and~\ref{table: attack_perplexity_qwen}, respectively, list perplexities of attacked watermarked texts at each attack probability ($\epsilon = 0.2$ or $\epsilon = 0.4$) under three language models. In terms of the perplexity metric, the superiority of addressing TI exists in attacked scenarios as well as unattacked scenarios (shown in Tables~\ref{table: watermark_llama_results},~\ref{table: watermark_swallow_results}, and~\ref{table: watermark_qwen_results}).

% \usepackage{multirow}

\begin{table*}[!t]
\small
\renewcommand{\arraystretch}{1.0}
\centering

  \scalebox{1.0}{
\begin{tabular}{l|l|rrr|rr|rr}
\toprule[1.0pt]
\multicolumn{2}{c|}{\multirow{2}{*}{Watermarking scheme}} & \multicolumn{3}{c|}{\textbf{Unattacked}}                                             & \multicolumn{2}{c|}{\textbf{Attacked} ($\epsilon = 0.2$)}  & \multicolumn{2}{c}{\textbf{Attacked} ($\epsilon = 0.4$)}                                        \\ \cline{3-9} 
\multicolumn{2}{c|}{}                                     & \begin{tabular}[c]{@{}r@{}}Watermark\\ Strength$\uparrow$\end{tabular} & AUROC$\uparrow$ & PPL$\downarrow$ & \begin{tabular}[c]{@{}r@{}}Watermark\\ Strength$\uparrow$\end{tabular} & AUROC$\uparrow$ & \begin{tabular}[c]{@{}r@{}}Watermark\\ Strength$\uparrow$\end{tabular} & AUROC$\uparrow$ \\ \midrule[1.0pt]
\multirow{2}{*}{\textbf{LeftHash}} & Original       & 8.05  &  0.999                                                     & 20.47 & \textbf{5.18} & 0.987 & \textbf{3.04} & 0.895   \\
                          & Post-hoc rollback & \textbf{8.07}  &    0.999                                                & \textbf{20.23} & 5.13 & \textbf{0.988} & 3.02 & \textbf{0.896}  \\ \hline
\multirow{2}{*}{\textbf{SelfHash}} & Original       & 7.96    &  0.998                                                   & 22.21 & 4.58 & 0.971 & 2.54 & 0.855   \\
                          & Post-hoc rollback & \textbf{7.99}   &  0.998                                                     & \textbf{21.72} & \textbf{4.67} & \textbf{0.975} & \textbf{2.58} & \textbf{0.859}    \\ \hline
\multirow{2}{*}{\textbf{Unigram}}  & Original       & 8.26   &  0.999                                                     & 19.80 & 6.98 & 0.988 & 5.54  & 0.917   \\
                          & Post-hoc rollback & \textbf{8.32}   &   0.999                                                   & \textbf{19.74} & \textbf{7.02} & 0.988 & \textbf{5.78}  & \textbf{0.921}   \\ \hline
\multirow{2}{*}{\textbf{Gumbel}}   & Original       & 29.97 &   \textbf{0.910}                                                        & 1.97 & 22.36 & 0.885 & 16.63 & 0.862   \\
                          & Post-hoc rollback & \textbf{31.24}  &  0.905                                                        & \textbf{1.95} & \textbf{23.33} & \textbf{0.887} & \textbf{17.01} &  \textbf{0.864}  \\
\bottomrule[1,0pt]
\end{tabular}}
 \caption{Quantitative comparison in various watermarking schemes on Swallow-7b and Japanese context.}
 \label{table: watermark_swallow_results}
\end{table*}

\begin{table*}[!t]
\small
\renewcommand{\arraystretch}{1.0}
\centering

  \scalebox{1.0}{
\begin{tabular}{l|l|rrr|rr|rr}
\toprule[1.0pt]
\multicolumn{2}{c|}{\multirow{2}{*}{Watermarking scheme}} & \multicolumn{3}{c|}{\textbf{Unattacked}}                                             & \multicolumn{2}{c|}{\textbf{Attacked} ($\epsilon = 0.2$)}  & \multicolumn{2}{c}{\textbf{Attacked} ($\epsilon = 0.4$)}                                        \\ \cline{3-9} 
\multicolumn{2}{c|}{}                                     & \begin{tabular}[c]{@{}r@{}}Watermark\\ Strength$\uparrow$\end{tabular} & AUROC$\uparrow$ & PPL$\downarrow$ & \begin{tabular}[c]{@{}r@{}}Watermark\\ Strength$\uparrow$\end{tabular} & AUROC$\uparrow$ & \begin{tabular}[c]{@{}r@{}}Watermark\\ Strength$\uparrow$\end{tabular} & AUROC$\uparrow$ \\ \midrule[1.0pt]
\multirow{2}{*}{\textbf{LeftHash}} & Original       & 8.47    & 0.999                                                        & 62.42 & 5.36 & 0.991 & 2.85 & 0.901   \\
                          & Post-hoc rollback & \textbf{8.54}     & 0.999                                                      & \textbf{58.12} & \textbf{5.37} & \textbf{0.992} & \textbf{3.00} & \textbf{0.910}  \\ \hline
\multirow{2}{*}{\textbf{SelfHash}} & Original       & 8.29   &  1.000                                                       & 58.00 & 4.67 & 0.975 & 2.13 & 0.829  \\
                          & Post-hoc rollback & \textbf{8.35}  &   1.000                                                      & \textbf{53.16} & \textbf{4.68} & \textbf{0.978} & \textbf{2.22} & \textbf{0.835}  \\ \hline
\multirow{2}{*}{\textbf{Unigram}}  & Original       & 9.23   &  1.000                                                     & 54.10 & 7.67  & 0.994 & \textbf{6.22} & \textbf{0.954}  \\
                          & Post-hoc rollback & \textbf{9.28}  &    1.000                                                     & \textbf{51.84} & \textbf{7.72} & 0.994 & 6.07 & 0.939  \\ \hline
\multirow{2}{*}{\textbf{Gumbel}}   & Original       & 22.09    &  0.897                                                      & 2.37 & 15.96 & \textbf{0.859} & 11.58 & 0.793  \\
                          & Post-hoc rollback & \textbf{22.46}   &  \textbf{0.900}                                                       & 2.37 & \textbf{16.27} & 0.858 & \textbf{12.17} & \textbf{0.796}  \\
\bottomrule[1,0pt]
\end{tabular}}
 \caption{Quantitative comparison in various watermarking schemes on Qwen2.5-7b and Chinese context.}
 \label{table: watermark_qwen_results}
\end{table*}

\section{Text Samples}
\subsection{Samples of Stegotexts}

\begin{table*}[!t]
\small
\centering

\begin{tabular}{l|l|l|r|r|r}
\toprule[1pt]
Model      & Prompt & Stegotext & BPT & PPL & KLD \\ \midrule[1pt]
Llama-2-7b & \begin{tabular}[c]{@{}l@{}}Finding a high quality\\ photographer for\\ your family portrait, \end{tabular}     & \begin{tabular}[c]{@{}l@{}}Finding a high quality photographer for your\\ family portrait, event or wedding is not as easy\\ as it sounds. A photography business can offer\\ photographers for hire that are affordable, but\\ this does not mean that you have to compromise\\ on quality. Here are more tips to help you find \\ the ultimate business that\end{tabular} & 2.33       & 6.23   & 0.024   \\ \hline
Llama-2-7b & \begin{tabular}[c]{@{}l@{}} I’ve never understood \\ the  whole points thing, \\ scholarships,    \end{tabular}  & \begin{tabular}[c]{@{}l@{}} I’ve never understood  the whole points thing, \\ scholarships, etc. Now it sounds like a full ride \\ (or nearly so). The deal is still a good one. That\\ probably sounds heartless to the \end{tabular}    & 4.41   & 26.64   & 0.009   \\ \hline
Swallow-7b &  \begin{CJK}{UTF8}{min} \begin{tabular}[c]{@{}l@{}}  該当する商品が、\\売り\end{tabular} \end{CJK} & \begin{CJK}{UTF8}{min} \begin{tabular}[c]{@{}l@{}} 該当する商品が、売りたい商品の条件に一\\致している場合に、お勧めできる相手を探\\して、1社から該当する会社への発送を依\\頼する方法が出品 \end{tabular} \end{CJK} & 3.76 & 21.99 & 0.059 \\ \hline
Swallow-7b & \begin{CJK}{UTF8}{min} \begin{tabular}[c]{@{}l@{}}褐色 仙台八幡、\\汚れ \end{tabular} \end{CJK} & \begin{CJK}{UTF8}{min} \begin{tabular}[c]{@{}l@{}} 褐色 仙台八幡、汚れ再生塗装 : 仙台 外壁\\塗装 お客さまの声
外壁塗装終わり、綺\\麗に汚れが落ちて大満足!仙台中央 \end{tabular} \end{CJK} & 4.27 & 71.62 & 0.013 \\ \hline
Qwen2.5-7b & \begin{CJK}{UTF8}{gbsn} \begin{tabular}[c]{@{}l@{}}  “全脑开发”真能让孩 \end{tabular} \end{CJK} & \begin{CJK}{UTF8}{gbsn} \begin{tabular}[c]{@{}l@{}}  “全脑开发”真能让孩成长吗？这个“脑”就是\\指孩子大脑的开发潜能。全脑开发的主要\\内容包括情绪、记忆力、思维力、想象力、\\创造力等智力因素和观察力、注意力、\\思维力等非智力因素的 \end{tabular} \end{CJK} & 2.72 & 11.39 & 0.096 \\ \hline
Qwen2.5-7b & \begin{CJK}{UTF8}{gbsn} \begin{tabular}[c]{@{}l@{}} 荷兰国际集团预计\\美国 \end{tabular} \end{CJK} & \begin{CJK}{UTF8}{gbsn} \begin{tabular}[c]{@{}l@{}} 荷兰国际集团预计美国五金商品经销商对\\抵押率为25\%的卷心菜/枫叶拱 \end{tabular} \end{CJK} & 7.53 & 233.92 & 0.027 \\
\bottomrule[1pt]
\end{tabular}
 \caption{Examples of stegotexts generated with our stepwise verification method.}
 \label{table: stegotext examples}
\end{table*}

\begin{table*}[!t]
\small
\centering

\begin{tabular}{l|l|l|l|r|r}
\toprule[1pt]
Model  & Method    & Prompt & Stegotext & \begin{tabular}[c]{@{}r@{}}Watermark\\ strength \end{tabular}  & PPL  \\ \midrule[1pt]
Llama-2-7b & LeftHash & \begin{tabular}[c]{@{}l@{}}In fact, there are\\ currently over a\\ million international \end{tabular}     & \begin{tabular}[c]{@{}l@{}}In fact, there are currently over a million \\ international students enrolled around  \\the world?  These valuable languages, \\ particularly  the most important ones, offer \\ infinite  possibilities; with them, you’ll \\ set out[...continues]
\end{tabular} & 8.19 & 16.79     \\ \hline
Llama-2-7b & SelfHash & \begin{tabular}[c]{@{}l@{}} So you have your\\ children writing every\\ day. Great!    \end{tabular}  & \begin{tabular}[c]{@{}l@{}} So you have your children writing every\\ day. Great! But it’s important not to think\\ “writing” means that only stories and\\ poems will qualify. In the course of life,\\ children will write in any number of\\ ways:[...continues]  \end{tabular}    & 6.82 & 16.55     \\ \hline
Swallow-7b &  LeftHash &  \begin{CJK}{UTF8}{min} \begin{tabular}[c]{@{}l@{}}  『秋晴れや 千種\\若水 \end{tabular} \end{CJK} & \begin{CJK}{UTF8}{min} \begin{tabular}[c]{@{}l@{}} 『秋晴れや 千種若水 父逝きぬ』 (\\『敬老の日』) 18年前に亡くした父\\親のことをのせた俳句です。2003年\\(平成15年)とイ「ンタネット」がリ\\1995年(平成7年)より普[...continues] \end{tabular} \end{CJK} &  9.71 & 32.14 \\ \hline
Swallow-7b & Unigram & \begin{CJK}{UTF8}{min} \begin{tabular}[c]{@{}l@{}}【札幌（新千歳）\\発】 \end{tabular} \end{CJK} & \begin{CJK}{UTF8}{min} \begin{tabular}[c]{@{}l@{}} 【札幌（新千歳）発】 北海道クルー\\ズ旅物語 こんな方にお勧め! (もちろ\\ん、リクエスト型をお選びいただい\\た方に限ります) 札幌観光と北海道を\\網羅する内容にしたい![...continues] \end{tabular} \end{CJK} & 7.50  & 14.36 \\ \hline
Qwen2.5-7b &  LeftHash & \begin{CJK}{UTF8}{gbsn} \begin{tabular}[c]{@{}l@{}}  书法教育：临摹\\还是创 \end{tabular} \end{CJK} & \begin{CJK}{UTF8}{gbsn} \begin{tabular}[c]{@{}l@{}} 书法教育：临摹还是创发我们的书法\\临写，可以从追溯300年传统书法产\\生之时就分为帖学和碑学两个主要\\的传统。行摹也是帖学书法传承的主\\要途径。但在现实生活[...continues]   \end{tabular} \end{CJK} & 8.55 & 116.27 \\ \hline
Qwen2.5-7b & Gumbel & \begin{CJK}{UTF8}{gbsn} \begin{tabular}[c]{@{}l@{}} 湖南茶博会全\\省30个 \end{tabular} \end{CJK} & \begin{CJK}{UTF8}{gbsn} \begin{tabular}[c]{@{}l@{}}  湖南茶博会全省30个市州组团来长\\沙设馆参展星辰在线5月15日讯5月\\15日，2023湖南茶叶博览会(简称“茶博\\会”)专业观众招募活动媒体吹\\风会在长沙举行，[...continues]\end{tabular} \end{CJK} & 4.48 & 2.76 \\
\bottomrule[1pt]
\end{tabular}
 \caption{Examples of watermarked generated with our post-hoc rollback method.}
 \label{table: watermark examples}
\end{table*}

\begin{table*}[!t]
\small
\centering
\begin{tabular}{l|r|r|r|r}
\toprule[1pt]
\multicolumn{5}{c}{\textbf{Llama-2-7b}} \\ \midrule[1pt]
\textbf{Generated token list}   & [Previous tokens]... & `eye'   & `q' & - \\
\textbf{Predicted probability} & [Previous probabilities]... & $6.59 \times 10^{-5}$ & $2.94 \times 10^{-4}$ & -\\ \hline
\textbf{Retokenized token list} & [Previous tokens]... & \cellcolor{softred} `e'  & \cellcolor{softred} `y'  & \cellcolor{softred} `eq' \\
\textbf{Predicted probability} & [Previous probabilities]... 
& \cellcolor{softred} $9.49 \times 10^{-6}$ & \cellcolor{softred} $9.79 \times 10^{-7}$ & \cellcolor{softred} $9.27 \times 10^{-8}$ \\ \midrule[1pt]

\multicolumn{5}{c}{\textbf{Swallow-7b}} \\ \midrule[1pt]
\textbf{Generated token list}   & [Previous tokens]... & \begin{CJK}{UTF8}{min} `に' \end{CJK}  & \begin{CJK}{UTF8}{min} `生' \end{CJK} & \begin{CJK}{UTF8}{min} `え' \end{CJK} \\
\textbf{Predicted probability} & [Previous probabilities]... & $1.91 \times 10^{-3}$ & $2.43 \times 10^{-4}$ & $4.53 \times 10^{-4}$ \\ \hline
\textbf{Retokenized token list} & [Previous tokens]... & \begin{CJK}{UTF8}{min} `に' \end{CJK} & \cellcolor{softred} \begin{CJK}{UTF8}{min} `生え' \end{CJK} & - \\
\textbf{Predicted probability} & [Previous probabilities]... & $1.91 \times 10^{-3}$ & \cellcolor{softred} $2.17 \times 10^{-6}$ & - \\ \midrule[1pt]

\multicolumn{5}{c}{\textbf{Qwen2.5-7b}} \\ \midrule[1pt]
\textbf{Generated token list}   & [Previous tokens]... & \begin{CJK}{UTF8}{gbsn} `医生' \end{CJK}  & \begin{CJK}{UTF8}{gbsn} `态度' \end{CJK} &  -  \\
\textbf{Predicted probability} & [Previous probabilities]... & $4.06 \times 10^{-3}$ & $5.27 \times 10^{-2}$ & - \\ \hline
\textbf{Retokenized token list} & [Previous tokens]... & \begin{CJK}{UTF8}{gbsn} \cellcolor{softred}`医' \end{CJK} & \cellcolor{softred} \begin{CJK}{UTF8}{gbsn} `生态'\end{CJK} & \cellcolor{softred} \begin{CJK}{UTF8}{gbsn} `度'\end{CJK} \\
\textbf{Predicted probability} & [Previous probabilities]... & \cellcolor{softred} $1.94 \times 10^{-2}$ & \cellcolor{softred}$1.36 \times 10^{-6}$ & \cellcolor{softred} $4.61 \times 10^{-6}$ \\ \bottomrule[1pt]

\end{tabular}
 \caption{Examples of how TI effects predicted probabilities of tokens.}
 \label{table: glitch-PPL examples}

\end{table*}

Table~\ref{table: stegotext examples} presents examples of stegotexts generated by our proposed stepwise verification method. Each generated text embeds a 128-bit random secret message. Following the approach of Ziegler et al.~\cite{ziegler-etal-2019-neural}, we terminate the generation process once the proposed method has finished embedding the message. 

\subsection{Samples of Watermarked Texts}
Table~\ref{table: watermark examples} lists some examples of watermarked texts generated by our proposed post-hoc rollback method, and as the number of generated tokens is 200 for each text, the complete watermarked texts are not provided due to space limitation.

\subsection{Samples of How TI Influences Perplexities}
\label{sec: samples of effects on perplexities}
Table~\ref{table: glitch-PPL examples} provides examples to clarify how consequential inconsistent tokens (CITs) affect the predicted probabilities of each token in the token list when the TI occurs, where CITs are marked in \colorbox{softred}{red background}. These examples show the fact that these predicted probabilities of CITs are generally lower than those of SITs, whereas only these CITs can be accessed by perplexity calculation. 

The expression of calculating perplexity is:
\[
\text{PPL} = \exp\left(-\frac{1}{N} \sum_{i=1}^{N} \log P(s^{(i)} \mid s^{(1):(i-1)}) \right)
\]
where $N$ is the length of the retokenized token list, $s^{(i)}$ denotes the $i^{th}$ token in this list, and $P(s^{(i)} \mid s^{(1):(i-1)})$ represents the predicted probability of $s^{(i)}$ according to historical $i-1$ tokens. Therefore, when the predicted probabilities of CITs are lower than SITs, the resulting perplexity is higher than that if perplexity calculation is based on the original generate token list (which includes SITs).

% \begin{figure}[!t]
%  \centering
%  \includegraphics[width=\columnwidth]{plot_topk_Llama.pdf} 
%  \caption{Average embedding capacities (bits per token, BPT) when using various disambiguation methods and our stepwise verification method, under Llama-2-7b and top-$k$ values.}
%  \label{fig: plot_topk_Llama}
% \end{figure}

% \begin{figure}[!t]
%  \centering
%  \includegraphics[width=\columnwidth]{plot_topk_Swallow.pdf} 
%  \caption{Average embedding capacities (bits per token, BPT) when using various disambiguation methods and our stepwise verification method, under Swallow-7b and top-$k$ values.}
%  \label{fig: plot_topk_Swallow}
% \end{figure}

% \begin{figure}[!t]
%  \centering
%  \includegraphics[width=\columnwidth]{plot_topk_Qwen.pdf} 
%  \caption{Average embedding capacities (bits per token, BPT) when using various disambiguation methods and our stepwise verification method, under Qwen2.5-7b and top-$k$ values.}
%  \label{fig: plot_topk_Qwen}
% \end{figure}

\begin{table}[!t]
\centering
\small
\begin{tabular}{l|rrr}
\toprule[1pt]
  & Llama-2-7b & Swallow-7b & Qwen2.5-7b \\
  \midrule[1pt]
$N=1$ &    $0.00\%$        &   $1.36\%$          &   $60.03\%$          \\
$N=2$ &         $3.61\%$    &      $68.02\%$       &    $27.69\%$         \\
$N=3$ &       $3.61\%$      &     $2.33\%$        &     $0.20\%$        \\
$...$ &     $...$        &      $...$       &    $...$         \\
$N=\infty$ &       $91.24\%$      &      $18.02\%$       &       $12.07\%$  \\
\bottomrule[1pt]
\end{tabular}
 \caption{The percentage that temporary inconsistency disappears after generating $N$ tokens afterwards.}
 \label{table: determine q}
\end{table}

\section{Determine Observation Period (q)}
\label{sec: determine q}
Table~\ref{table: determine q} lists the percentage that temporary inconsistency naturally disappears after generating $N$ tokens afterwards. And $N = \infty$ denotes a permanent inconsistency (in 100-token texts, 1000 samples for each model). According to this table, we can find that, in Swallow-7b and Qwen2.5-7b, most temporary inconsistencies disappear after generating 2 subsequent tokens (because of partial UTF-8 tokens), while in Llama-2-7b, most temporary inconsistencies are much more stable (because of special tokens).

Therefore, back to the principled way to determine, in Swallow-7b and Qwen2.5-7b, is set at least greater than 2 (we set $q=10$ in experiments) to avoid most temporary inconsistencies (to avoid the false positive), and in Llama-2-7b, can be set as a much smaller value (we set $q=2$ in experiments), because most inconsistencies which have happened will not disappear, which means that it is suitable to be fixed immediately.

Table~\ref{table: determine q} also explains the significant differences in the temporariness of inconsistent tokens between Llama-2-7b and the other two language models, as more than 90\% temporary inconsistencies do not disappear in Llama-2-7b (shown in Table~\ref{table:temporary_glitch_rate}).

\section{Efficiency of the Post-Hoc Rollback Method for Watermarking}

Table~\ref{table: watermark_efficiency} reports the running time (in seconds) for various watermarking schemes across different language models, based on 500 samples with 200 tokens each.
The results show that the runtime differences between the original methods and our proposed post-hoc rollback method are minimal (generally under 5\%) indicating that our approach introduces minor computational overhead.

% \usepackage{multirow}
\begin{table*}[!t]
\centering
\begin{tabular}{l|l|r|r|r}
\toprule[1.0pt]
\multicolumn{2}{l|}{Watermarking scheme}       & Llama-2-7b & Swallow-7b & Qwen2.5-7b \\
\midrule[1.0pt]
\multirow{2}{*}{\textbf{LeftHash}} & Original          &         6.89   &    7.13        &    6.03        \\
                          & Post-hoc rollback & $_{(+1.16\%)}$   6.97          &  $_{(+3.51\%)}$ 7.38          &  $_{(+3.98\%)}$  6.27        \\ \hline
\multirow{2}{*}{\textbf{SelfHash}} & Original          & 7.04           &  7.24          &    5.96        \\
                          & Post-hoc rollback &  $_{(+0.85\%)}$ 7.10          & $_{(+3.89\%)}$  7.52         &  $_{(+2.34\%)}$  6.10        \\ \hline
\multirow{2}{*}{\textbf{Unigram}}  & Original          &  6.88          &     7.17       &  6.11          \\
                          & Post-hoc rollback &    $_{(+2.90\%)}$  7.08        &  $_{(+5.44\%)}$   7.56       & $_{(+5.24\%)}$  6.43         \\ \hline
\multirow{2}{*}{\textbf{Gumbel}}   & Original          &   6.85         &    7.26        & 6.00           \\
                          & Post-hoc rollback &   $_{(+2.04\%)}$ 6.99         & $_{(+3.03\%)}$   7.48        & $_{(+4.17\%)}$ 6.25 \\
\bottomrule[1.0pt]
\end{tabular}
 \caption{Running time (seconds) in various watermarking schemes when different language models are adopted.}
 \label{table: watermark_efficiency}
\end{table*}

\begin{table*}[!t]
\centering
\small
\begin{tabular}{l|l|rr|rr|rr}
\toprule[1pt]
\multicolumn{2}{l|}{\multirow{2}{*}{Watermarking scheme}} & \multicolumn{2}{c|}{Llama-2-7b}                                        & \multicolumn{2}{c|}{Swallow-7b}                                        & \multicolumn{2}{c}{Qwen2.5-7b}                                        \\ \cline{3-8}
\multicolumn{2}{l|}{}                                     & \begin{tabular}[c]{@{}r@{}}Watermark \\ Strength $\uparrow$\end{tabular} & AUROC $\uparrow$ & \begin{tabular}[c]{@{}r@{}}Watermark \\ Strength $\uparrow$ \end{tabular} & AUROC $\uparrow$ & \begin{tabular}[c]{@{}r@{}}Watermark \\ Strength $\uparrow$\end{tabular} & AUROC $\uparrow$ \\ \midrule[1pt]
\multirow{2}{*}{\textbf{LeftHash}}                        & Original               &        2.57        &  0.876     &  \textbf{2.92}   &  \textbf{0.887}     &  3.47  &   0.913    \\
                                & Post-hoc rollback      &   \textbf{2.60}     &  \textbf{0.877}     &    2.91       &  0.886     & 3.47  & \textbf{0.915}     \\ \hline
\multirow{2}{*}{\textbf{SelfHash}}       & Original      & 1.86  &   0.803   &  2.35     &  0.837       & 2.42  &  0.861  \\
                                & Post-hoc rollback      &  \textbf{1.99}    &  \textbf{0.815}    &    2.35      &     \textbf{0.840}   &   \textbf{2.44}   &    \textbf{0.865}   \\  \hline
\multirow{2}{*}{\textbf{Unigram}}        & Original    & 4.56  &   0.898    &   6.63  &   0.946    &     4.39       &   0.918    \\
                                & Post-hoc rollback      &     \textbf{4.59}      &   \textbf{0.905}    &   \textbf{6.69}       &  \textbf{0.950}     &    \textbf{4.51} &    \textbf{0.923}   \\  \hline
\multirow{2}{*}{\textbf{Gumbel}}         & Original               & 10.62 & 0.887
      &    18.18       &  0.877     & 15.00  &   \textbf{0.851}    \\
                                & Post-hoc rollback      &    \textbf{11.07}        &   \textbf{0.895}    &   \textbf{18.56}                            & \textbf{0.879}     &  \textbf{15.32}        &  0.850   \\ 
                                \bottomrule[1pt]
\end{tabular}
 \caption{Quantitative comparison in various watermarking schemes under GPT-4o paraphrasing attack when different language models are adopted.}
 \label{table: GPT4 attack}
\end{table*}

\section{Attacking Watermarks with GPT-4o Paraphrasing}
\label{sec: GPT-4o Attack}
Table~\ref{table: GPT4 attack} lists the average experimental statistics where the watermarked texts are attacked by GPT-4o paraphrasing (\texttt{temperature = 1} and \texttt{max\_completion\_tokens = 2048}). Specifically, the prompt template is:
\begin{tcolorbox}
[colback=gray!10!white, colframe=gray!80!black, title= \texttt{Paraphrasing template of GPT-4o}]
    % \small
    \textbf{System message:} You are tasked to paraphrase. Please directly paraphrase the text you receive (in the corresponding language).
    
    \textbf{User prompt:} <text>
\end{tcolorbox}

From Table~\ref{table: GPT4 attack}, we can find that our proposed post-hoc rollback method for watermarking overall improves the robustness against paraphrasing attacks based on strong a large language model.

\end{document}